\documentclass[10pt,twocolumn,letterpaper]{article}

\usepackage[pagenumbers]{wacv} 

\usepackage{booktabs}
\usepackage{multirow}
\usepackage[table]{xcolor}
\usepackage{xcolor}
\usepackage{colortbl}
\usepackage{array}
\usepackage[most]{tcolorbox}
\usepackage{adjustbox}

\definecolor{wacvblue}{rgb}{0.21,0.49,0.74}
\usepackage[pagebackref,breaklinks,colorlinks,allcolors=wacvblue]{hyperref}

\definecolor{glascolor}{RGB}{239,246,252}
\definecolor{ph2color}{RGB}{247,241,249}
\definecolor{hmepscolor}{RGB}{239,248,242}
\definecolor{ourscolor}{RGB}{246,246,246}

\def\wacvPaperID{1367} 
\def\confName{WACV}
\def\confYear{2027}

\title{Semi-Supervised Biomedical Image Segmentation via Diffusion Models and Teacher-Student Co-Training}

\author{Luca Ciampi,
Gabriele Lagani,
Giuseppe Amato, 
Fabrizio Falchi \\ [0.3em]
Institute of Information Science and Technologies (ISTI-CNR), Pisa, Italy \\  [0.3em]
{\tt\small \{luca.ciampi\}@isti.cnr.it}}

\begin{document}
\maketitle
\begin{abstract}
Supervised deep learning achieves strong performance in biomedical image segmentation but relies on costly pixel-wise annotations, motivating semi-supervised approaches that exploit unlabeled data. We introduce a diffusion-based teacher--student framework in which segmentation predictions are used to condition image denoising, encouraging the production of more informative pseudo-labels. The teacher is first pretrained through an unsupervised reconstruction task using diffusion-style corruption, timestep conditioning, and denoising. Starting from a corrupted empty mask, the model predicts an intermediate segmentation that conditions image denoising, encouraging the predicted mask to capture structural information useful for recovering the original image. The resulting teacher is then co-trained with a student using supervised segmentation on labeled samples and cross pseudo-supervision on unlabeled data. We further introduce a multi-round extension during co-training, in which the teacher generates multiple stochastic image reconstructions and corresponding segmentation predictions, providing additional reconstruction and alignment signals to improve its pseudo-labels. We evaluate the proposed framework on three public 2D biomedical segmentation datasets and a 3D left atrial segmentation benchmark. Across several labeling regimes, our method achieves competitive or superior performance compared with state-of-the-art semi-supervised approaches, with the largest gains observed under severe label scarcity.
\end{abstract}

\section{Introduction}
\label{sec:intro}

Deep learning (DL) models have demonstrated strong performance in biomedical image segmentation~\cite{10.1007/978-3-319-24574-4_28,DBLP:journals/corr/ChenPSA17,7298965,7785132,10.1007/978-3-030-87199-4_16,9706678,10.1007/978-3-031-25066-89,hebbian_eccv_workshop,DBLP:journals/nn/CiampiLAF26}. However, these methods typically rely on large amounts of annotated training data for supervised learning. In the biomedical domain, this requirement represents a particularly severe bottleneck, as pixel-wise annotation often requires substantial time from highly specialized domain experts and may also be affected by inter-observer variability. This time-consuming and resource-intensive process significantly limits the applicability of such methods in large-scale clinical settings~\cite{9625988,Isensee_2020,10.1007/978-3-319-46723-8_49,10.1007/978-3-030-87193-2_4,10.1016/j.patcog.2022.108673,10332179}.

To address this limitation, semi-supervised learning combines supervised training on a limited set of labeled samples with additional learning from a larger pool of unlabeled data~\cite{DBLP:conf/nips/TarvainenV17,DBLP:conf/cvpr/IscenTAC19,DBLP:journals/nn/VermaKLKSBL22,chen2020b,DBLP:journals/artmed/LeeHWHLHT24,DBLP:journals/artmed/MeiYZZCYW24,DBLP:journals/artmed/LiZYZWZLSW24}. Among semi-supervised approaches, teacher--student methods have proven particularly effective~\cite{DBLP:conf/nips/TarvainenV17,10.1007/978-3-030-32245-8_67,9577639}. In these methods, a teacher model provides supervisory signals, typically in the form of pseudo-labels or prediction targets, to guide learning from unlabeled data. A key limitation of pseudo-label-based learning, however, is confirmation bias, whereby inaccurate predictions may be reinforced and accumulate errors during training~\cite{DBLP:conf/eccv/OuyangBCKQR20,lee2013}. Thus, the quality of supervisory predictions is critical. In semantic segmentation, this is particularly challenging, as predictions must not only assign the correct semantic classes but also preserve accurate boundaries of the segmented structures~\cite{10376766,LUO2022102517,luo2021}.

\begin{figure*}
\centering
\includegraphics[width=0.9\linewidth]{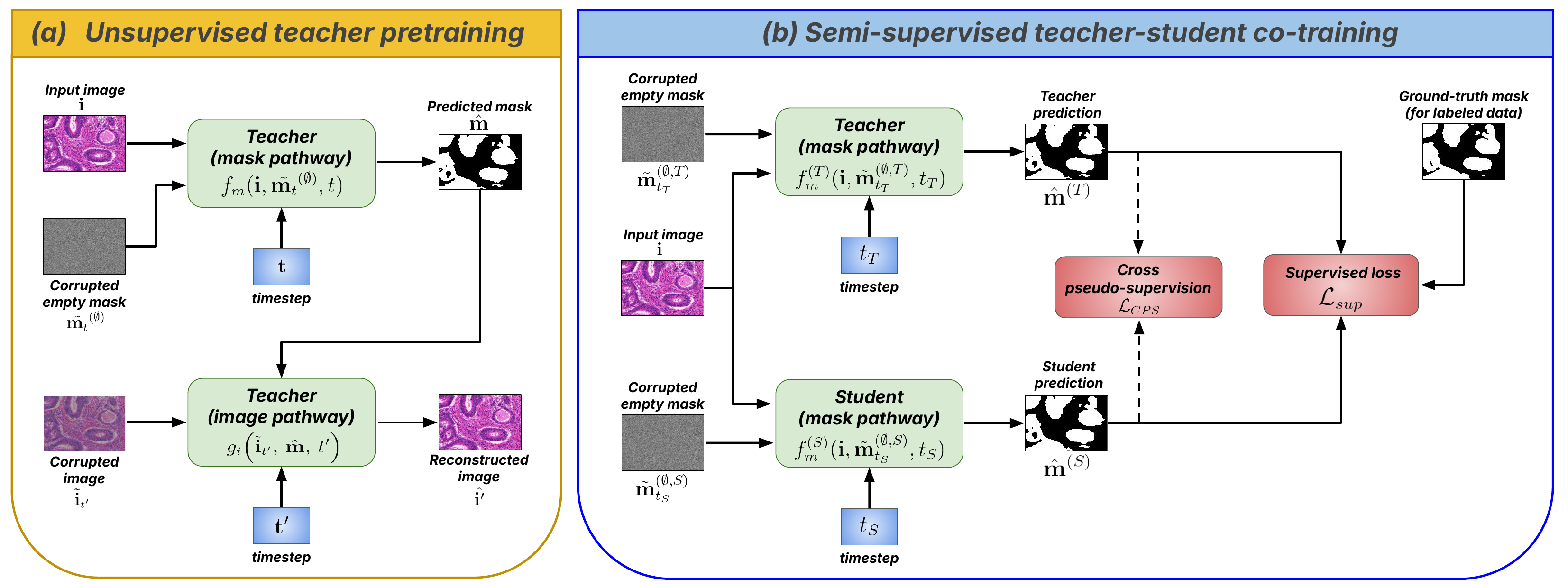}
\caption{\textbf{Overview of the main training procedure of the proposed teacher--student framework for semi-supervised biomedical image segmentation.} (a) Unsupervised teacher pretraining. The teacher mask pathway receives the input image together with a noise-corrupted empty segmentation mask and produces an intermediate segmentation prediction. This prediction conditions the teacher image pathway, which processes an independently corrupted version of the input image to reconstruct a clean-image estimate. (b) Semi-supervised teacher--student co-training. The teacher and student mask pathways process the same input image together with independently corrupted empty masks and independently sampled diffusion timesteps, producing the corresponding segmentation predictions. For labeled samples, both predictions are supervised by the ground-truth mask through the supervised segmentation loss. For unlabeled samples, cross pseudo-supervision (CPS) enables the prediction of each model to provide pseudo-supervision to the other.}
\label{fig:teaser}
\end{figure*}

In this paper, we introduce a novel semi-supervised approach to biomedical image segmentation that follows a teacher--student co-training paradigm. Our formulation adopts the forward diffusion process, timestep conditioning, and denoising objectives used in denoising diffusion probabilistic models (DDPMs)~\cite{DBLP:conf/nips/HoJA20}. The core intuition is that, by coupling a segmentation prediction task with a conditional image-denoising objective, the teacher is encouraged to generate an intermediate mask that retains structural information useful for recovering the original image from its corrupted version, thereby improving its suitability as a pseudo-label.
Specifically, the proposed framework combines unsupervised teacher pretraining with semi-supervised teacher--student co-training and further incorporates a multi-round extension during co-training. As illustrated in Fig.~\ref{fig:teaser}(a), the teacher first undergoes an unsupervised reconstruction-based pretraining stage using only unlabeled images. Given an input image together with a corrupted empty mask and the corresponding diffusion timestep, the teacher generates an intermediate segmentation prediction, which is then used to condition the denoising of an independently corrupted version of the same image. This objective encourages the segmentation output to preserve information relevant to the input image and prepares the teacher to produce informative pseudo-labels. During the subsequent co-training stage (Fig.~\ref{fig:teaser}(b)), the pretrained teacher and the student are jointly optimized. Both models receive the same input image together with independently corrupted empty masks and independently sampled diffusion timesteps and produce the corresponding segmentation predictions. Ground-truth masks supervise both models on labeled samples, whereas cross pseudo-supervision enables them to learn from each other's predictions on unlabeled samples. Finally, in the multi-round extension, the teacher generates multiple stochastic image reconstructions and corresponding segmentation predictions, which provide additional reconstruction and alignment signals to improve its pseudo-labels during co-training.

We evaluate our methodology on three widely used public datasets for 2D medical image segmentation~\cite{9625988,10.1007/978-3-030-87193-2_4,10.1016/j.patcog.2022.108673,10332179}, covering different imaging modalities and segmentation tasks. Specifically, we assess our approach on GlaS~\cite{SIRINUKUNWATTANA2017489} for colorectal gland segmentation in Hematoxylin and Eosin (H\&E)-stained histological images, PH2~\cite{6610779} for skin lesion segmentation in dermoscopic images, and HMEPS~\cite{raffaele_mazziotti_2021_4488164} for pupil segmentation in grayscale eye images. Additionally, we extend our evaluation to volumetric data using the LA~\cite{XIONG2021101832} dataset for left atrial segmentation in MRIs. We benchmark our method against several state-of-the-art semi-supervised approaches, including methods based on pseudo-labeling, consistency regularization, and entropy minimization~\cite{DBLP:conf/cvpr/VuJBCP19,10.1007/978-3-030-32245-8_67,9157032,9577639,LUO2022102517,Luo_2021}, as well as two-stage pipelines that employ recent self-supervised learning (SSL) paradigms during the initial unsupervised pretraining phase~\cite{10.1007/978-3-030-58526-6_45,perez2025}. Our results show that the proposed approach achieves competitive performance across different levels of label scarcity and reaches state-of-the-art performance in several experimental settings.

Concretely, our contributions are as follows:
\begin{itemize}

\item We propose a novel semi-supervised teacher--student framework for biomedical image segmentation that integrates the forward diffusion process, timestep conditioning, and denoising objectives used in DDPMs. The teacher is first trained on unlabeled images through a reconstruction-based pretraining objective that couples segmentation prediction with conditional image denoising, and is subsequently co-trained with a student using ground-truth supervision on labeled samples and cross pseudo-supervision on unlabeled samples.

\item We introduce a multi-round extension in which the teacher generates multiple stochastic image reconstructions and corresponding segmentation predictions, providing additional reconstruction and alignment signals to improve its pseudo-labels during co-training.

\item We evaluate the proposed framework on three public 2D biomedical image segmentation benchmarks across different imaging modalities and levels of label availability, and further evaluate it on a 3D left atrial segmentation dataset. We also conduct ablation studies to assess the contribution of the main components and configurations of the proposed framework.

\end{itemize}

\section{Related Work}  
\label{sec:related}
Deep learning has led to substantial advances in semantic segmentation, with architectures such as FCN~\cite{7298965}, SegNet~\cite{7803544}, and DeepLabV3~\cite{DBLP:journals/corr/ChenPSA17}, while U-Net-like models have become particularly widespread in biomedical imaging~\cite{10.1007/978-3-319-46723-8_49,7785132,9706678,10.1007/978-3-031-25066-89,Isensee_2020,CIAMPI2022102500,DBLP:conf/visapp/CiampiCAG22}. However, the scarcity of labeled biomedical data has motivated label-efficient learning strategies that exploit unlabeled samples alongside limited annotations~\cite{bengio2007,larochelle2009}. 

\paragraph{Pseudo-labeling and teacher--student approaches.}
Pseudo-labeling methods exploit model predictions as supervisory targets for unlabeled samples. These predictions may be generated by the same model or by a separate network, and can substantially improve segmentation performance when sufficiently accurate~\cite{DBLP:conf/nips/TarvainenV17,DBLP:conf/cvpr/IscenTAC19}. The Mean Teacher (MT) framework~\cite{DBLP:conf/nips/TarvainenV17} employs teacher and student models with the same architecture and updates the teacher parameters as an exponential moving average of the student parameters. The uncertainty-aware Mean Teacher (UAMT) method~\cite{10.1007/978-3-030-32245-8_67} adapts this strategy to biomedical image segmentation by incorporating uncertainty estimation. Cross pseudo-supervision (CPS)~\cite{9577639}, instead, employs two independently initialized networks that supervise each other using their predicted label maps. Our framework builds on the mutual-supervision principle of CPS, while introducing an unsupervised reconstruction-based pretraining stage for the teacher and a multi-round mechanism that provides additional reconstruction and alignment signals during co-training.

\paragraph{Consistency-, entropy-, and adversarial methods.}
Consistency-based semi-supervised methods exploit unlabeled samples by encouraging compatible predictions under different perturbations of the same input or representation~\cite{DBLP:conf/cvpr/IscenTAC19,DBLP:conf/nips/SohnBCZZRCKL20}. Cross-consistency training (CCT)~\cite{9157032} perturbs latent representations and enforces agreement among the resulting segmentation outputs. Uncertainty-rectified pyramid consistency (URPC)~\cite{LUO2022102517} instead promotes consistency among predictions produced at different decoder levels, while DTC~\cite{Luo_2021} jointly predicts a pixel-wise segmentation map and a geometry-aware level-set representation and enforces consistency between the two task outputs. Unlabeled data can also be exploited through entropy-based objectives. Vu et al.~\cite{DBLP:conf/cvpr/VuJBCP19} introduced direct entropy minimization and adversarial entropy alignment for unsupervised domain adaptation; the former can also be used in semi-supervised segmentation to encourage confident predictions on unlabeled images.

\paragraph{Two-stage label-efficient approaches.}
An alternative label-efficient strategy separates the use of unlabeled and labeled data into two stages: an initial unsupervised or self-supervised pretraining phase followed by adaptation to the target task using a limited set of annotations~\cite{bengio2007,larochelle2009,kingma2014b,zhang2016}. The method proposed in~\cite{10.1007/978-3-030-58526-6_45}, for example, learns representations using superpixel-derived pseudo-labels. More recently, large-scale biomedical encoders such as RAD-DINO~\cite{perez2025}, UNI~\cite{chen2024}, and Prov-GigaPath~\cite{xu2024} have leveraged large-scale self-supervised DINO-based pretraining~\cite{DBLP:conf/iccv/CaronTMJMBJ21} to learn transferable representations for downstream tasks. Unlike these two-stage approaches, our framework does not restrict the use of unlabeled data to pretraining: after unsupervised teacher pretraining, teacher and student remain jointly optimized using ground-truth supervision on labeled samples and cross pseudo-supervision on unlabeled samples.

\paragraph{Diffusion-based segmentation methods.}
Denoising diffusion models have recently been applied to biomedical image segmentation by formulating mask prediction as an image-conditioned denoising process. MedSegDiff~\cite{DBLP:conf/midl/WuFFZYXL023} and MedSegDiff-V2~\cite{DBLP:conf/aaai/Wu0FXJ024} progressively recover segmentation maps through iterative reverse diffusion, whereas BerDiff~\cite{DBLP:conf/miccai/ChenWS23} formulates the diffusion process in the Bernoulli domain to better accommodate discrete binary masks and exploits stochastic sampling to generate diverse segmentation outputs. These approaches primarily use diffusion as the segmentation model itself, commonly within fully supervised or probabilistic segmentation settings. In contrast, our framework employs forward diffusion, timestep conditioning, and conditional denoising objectives to support pseudo-label generation within a semi-supervised teacher--student setting, rather than relying on iterative reverse diffusion as its primary segmentation inference mechanism.

\begin{figure}[!t]
    \centering
    \includegraphics[width=\linewidth]{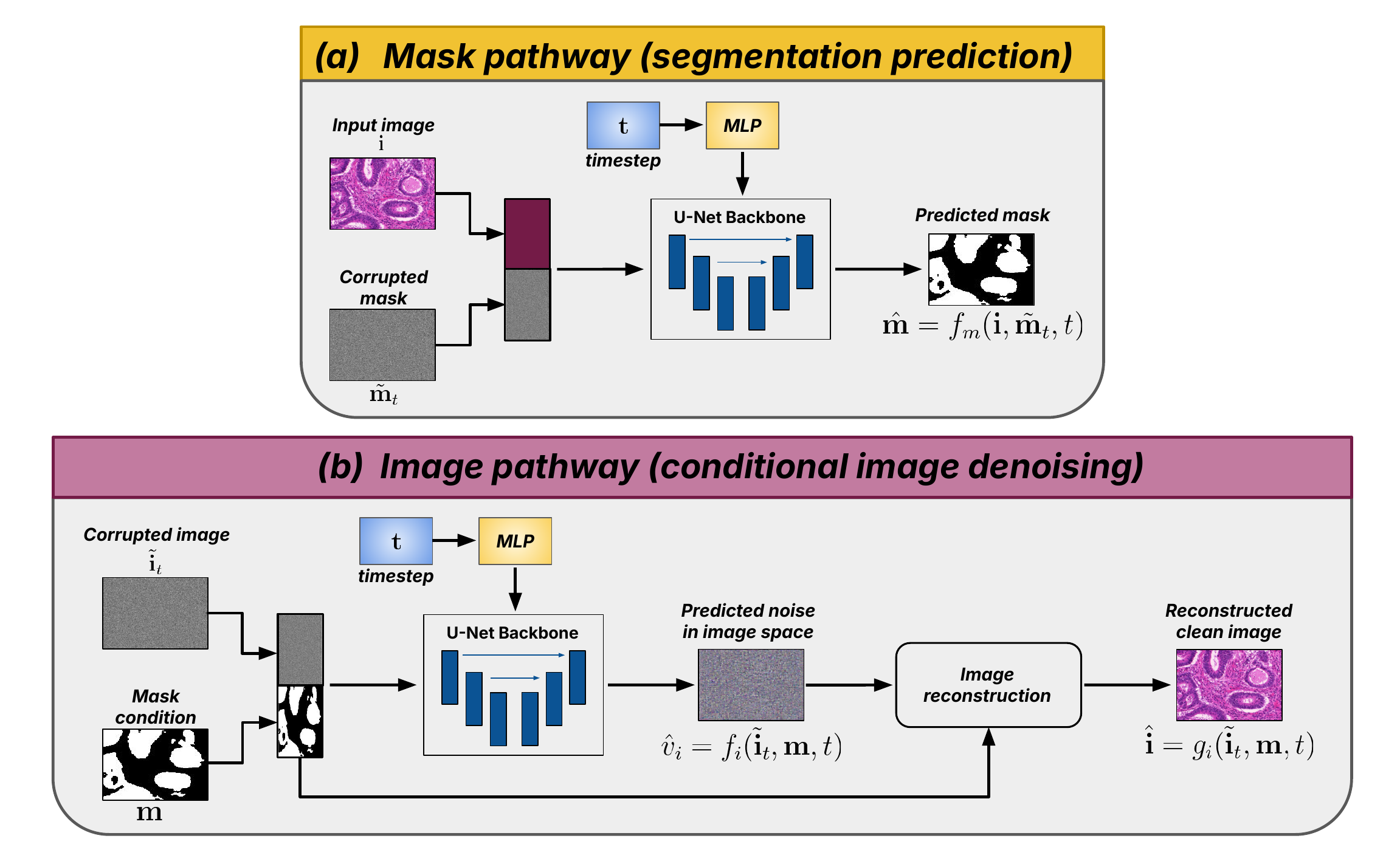}
    \caption{\textbf{Mask and image computational pathways.} The mask pathway predicts a segmentation mask from a clean image, a corrupted mask, and the diffusion timestep. The image pathway predicts the noise added to a corrupted image conditioned on a mask, and the corresponding clean-image estimate is then recovered through the reconstruction operator. During teacher pretraining, the predicted mask from the first pathway conditions the image pathway, as illustrated in Fig.~\ref{fig:teaser}.}
    \label{fig:image_mask_pathway}
\end{figure}

\section{Method}
\label{sec:method}

Our approach to semi-supervised biomedical image segmentation follows a teacher--student co-training framework and adopts the forward diffusion process, timestep conditioning, and denoising objectives used in denoising diffusion probabilistic models (DDPMs)~\cite{DBLP:conf/nips/HoJA20}. The framework combines unsupervised reconstruction-based teacher pretraining with semi-supervised teacher--student co-training and further incorporates a multi-round extension during co-training. During pretraining, an intermediate segmentation prediction generated by the teacher is used to condition the denoising of a corrupted version of the input image. During co-training, ground-truth masks supervise teacher and student predictions on labeled samples, whereas cross pseudo-supervision is used on unlabeled samples. The multi-round extension additionally generates stochastic reconstruction--segmentation pairs during co-training, providing further reconstruction and alignment signals to improve the teacher's pseudo-labels.

\subsection{Unsupervised Reconstruction-based Teacher Pretraining}
\label{sec:teacher_pretraining}

The teacher contains two distinct U-Net-based computational pathways: (i) a \textit{mask pathway}, which operates in the segmentation-mask domain, and (ii) an \textit{image pathway}, which operates in the image domain, as illustrated in Fig.~\ref{fig:image_mask_pathway}. Both pathways receive inputs corrupted according to the forward diffusion process used in DDPMs. For a normalized image $\mathbf{i}$ and a normalized $C$-channel mask representation $\mathbf{m}$, their corrupted versions at timestep $t$ are

\begin{equation}
\begin{aligned}
\tilde{\mathbf{i}}_t
&=
\sqrt{\bar{\alpha}_t}\,\mathbf{i}
+
\sqrt{1-\bar{\alpha}_t}\,\boldsymbol{\nu}_i,
\\
\tilde{\mathbf{m}}_t
&=
\sqrt{\bar{\alpha}_t}\,\mathbf{m}
+
\sqrt{1-\bar{\alpha}_t}\,\boldsymbol{\nu}_m,
\end{aligned}
\label{eq:forward_corruption}
\end{equation}

\noindent
where $\boldsymbol{\nu}_i$ and $\boldsymbol{\nu}_m$ are independent standard Gaussian tensors, $\boldsymbol{\nu}_i,\boldsymbol{\nu}_m\sim\mathcal{N}(\mathbf{0},\mathbf{I})$. The coefficient $\bar{\alpha}_t\in(0,1]$ is the cumulative signal-retention coefficient associated with the predefined noise schedule, with larger timesteps corresponding to stronger corruption. For each sample, the timestep is drawn from $t\sim\mathrm{Unif}\{0,\ldots,T-1\}$, where $T$ is the total number of diffusion timesteps.

The two pathways use different prediction objectives. Given a corrupted mask-domain input and a clean conditioning image, the mask pathway directly predicts a clean mask-domain output:

\begin{equation}
\hat{\mathbf{m}}
=
f_m
\left(
    \mathbf{i},
    \tilde{\mathbf{m}}_t,
    t
\right).
\label{eq:generic_mask_pathway}
\end{equation}

\noindent
Conversely, given a corrupted image and an uncorrupted mask-domain conditioning tensor, the image pathway predicts the Gaussian noise added to the image:

\begin{equation}
\hat{\boldsymbol{\nu}}_i
=
f_i
\left(
    \tilde{\mathbf{i}}_t,
    \mathbf{m},
    t
\right).
\label{eq:generic_image_pathway}
\end{equation}

\noindent
Both pathways are conditioned on the sampled timestep through a sinusoidal embedding processed by a multi-layer perceptron and incorporated into the internal feature representation. Each pathway produces its prediction in a single forward pass at the sampled timestep; no iterative reverse-diffusion chain is used.

Given the predicted noise, the corresponding clean-image estimate is

\begin{equation}
\hat{\mathbf{i}}
=
\sqrt{\frac{1}{\bar{\alpha}_t}}\,
\tilde{\mathbf{i}}_t
-
\sqrt{\frac{1}{\bar{\alpha}_t}-1}\,
\hat{\boldsymbol{\nu}}_i.
\label{eq:image_reconstruction}
\end{equation}

\noindent
For compactness, we define the corresponding image-reconstruction operator

\begin{equation}
\begin{aligned}
g_i
\left(
    \mathbf{x},
    \mathbf{m},
    t
\right)
&=
\sqrt{\frac{1}{\bar{\alpha}_t}}\,\mathbf{x}
\\
&\quad-
\sqrt{\frac{1}{\bar{\alpha}_t}-1}\,
f_i
\left(
    \mathbf{x},
    \mathbf{m},
    t
\right).
\end{aligned}
\label{eq:image_operator}
\end{equation}

\noindent
Hence, for $\mathbf{x}=\tilde{\mathbf{i}}_t$, the clean-image estimate is $\hat{\mathbf{i}}=g_i(\tilde{\mathbf{i}}_t,\mathbf{m},t)$.

Both pathways are conditioned on the sampled timestep through a sinusoidal
embedding processed by a multi-layer perceptron and incorporated into the
internal feature representation. Each pathway produces its prediction in a
single forward pass at the sampled timestep; no iterative reverse-diffusion
chain is used.

During teacher pretraining (Fig.~\ref{fig:teaser}(a)), the mask pathway receives a corrupted empty mask instead of an annotated segmentation mask. Let $\mathbf{m}^{(\emptyset)}$ denote an all-background one-hot mask. At a randomly sampled timestep $t$, this mask is corrupted as

\begin{equation}
\tilde{\mathbf{m}}^{(\emptyset)}_t
=
\sqrt{\bar{\alpha}_t}\,
\mathbf{m}^{(\emptyset)}
+
\sqrt{1-\bar{\alpha}_t}\,
\boldsymbol{\nu}_m.
\label{eq:empty_mask_corruption}
\end{equation}

\noindent
The mask pathway then produces an intermediate $C$-channel mask-domain prediction conditioned on the clean image:

\begin{equation}
\hat{\mathbf{m}}
=
f_m
\left(
    \mathbf{i},
    \tilde{\mathbf{m}}^{(\emptyset)}_t,
    t
\right).
\label{eq:pretraining_mask}
\end{equation}

\noindent
An independent timestep $t'$ and Gaussian tensor $\boldsymbol{\nu}'_i$ are then sampled, and the original image is independently corrupted as

\begin{equation}
\tilde{\mathbf{i}}_{t'}
=
\sqrt{\bar{\alpha}_{t'}}\,\mathbf{i}
+
\sqrt{1-\bar{\alpha}_{t'}}\,\boldsymbol{\nu}'_i,
\;
t'\sim\mathrm{Unif}\{0,\ldots,T-1\}.
\label{eq:pretraining_image_corruption}
\end{equation}

\noindent
Conditioned on the intermediate mask prediction, the image pathway estimates the noise added to the image:

\begin{equation}
\hat{\boldsymbol{\nu}}'_i
=
f_i
\left(
    \tilde{\mathbf{i}}_{t'},
    \hat{\mathbf{m}},
    t'
\right),
\label{eq:pretraining_noise_prediction}
\end{equation}

\noindent
and reconstructs the corresponding clean-image estimate as

\begin{equation}
\hat{\mathbf{i}}'
=
g_i
\left(
    \tilde{\mathbf{i}}_{t'},
    \hat{\mathbf{m}},
    t'
\right).
\label{eq:pretraining_image_reconstruction}
\end{equation}

\noindent
The reconstruction branch is optimized using the DDPM noise-prediction objective

\begin{equation}
\label{eq:cycle_consistency_loss}
\mathcal{L}_{\mathrm{rec}}^{(\mathrm{pre})}
=
\operatorname{MSE}
\left(
    \boldsymbol{\nu}'_i,
    \hat{\boldsymbol{\nu}}'_i
\right).
\end{equation}

Because $\hat{\boldsymbol{\nu}}'_i$ is predicted conditionally on $\hat{\mathbf{m}}$, gradients from Eq.~\ref{eq:cycle_consistency_loss} propagate through both computational pathways. The reconstruction objective therefore encourages the intermediate mask-domain output to retain information about the input image. No segmentation annotations are used during teacher pretraining. Once pretraining is completed, the resulting parameters initialize the teacher used in the subsequent co-training process.

\subsection{Semi-supervised Teacher--Student Co-training}
\label{sec:co_training}

The co-training stage uses the mask pathways of the teacher and student (Fig.~\ref{fig:teaser}(b)). Given an input image $\mathbf{i}$, each model receives an independently corrupted empty-mask initialization at a sampled timestep and produces a segmentation prediction:

\begin{equation}
\hat{\mathbf{m}}^{(k)}
=
f_m^{(k)}
\left(
    \mathbf{i},
    \tilde{\mathbf{m}}^{(\emptyset,k)}_{t_k},
    t_k
\right),
\
k\in\{\mathrm{Teach.},\mathrm{Stud.}\}.
\label{eq:teacher_student_predictions}
\end{equation}

\noindent
Each output is a $C$-channel segmentation score map containing one class score at each pixel. A hard segmentation mask is obtained by applying $\operatorname*{arg\,max}_{c}$ along the class dimension.

Let $\ell_{\mathrm{seg}}$ denote the pixel-wise segmentation loss used for co-training. For unlabeled samples, cross pseudo-supervision is defined as

\begin{equation}
\label{eq:loss-cps}
\begin{aligned}
\mathcal{L}_{\mathrm{CPS}}
={}&
\ell_{\mathrm{seg}}
\left(
    \hat{\mathbf{m}}^{(\mathrm{student})},
    \operatorname*{arg\,max}_{c}
    \hat{\mathbf{m}}^{(\mathrm{teacher})}
\right)
\\
&+
\ell_{\mathrm{seg}}
\left(
    \hat{\mathbf{m}}^{(\mathrm{teacher})},
    \operatorname*{arg\,max}_{c}
    \hat{\mathbf{m}}^{(\mathrm{student})}
\right).
\end{aligned}
\end{equation}

\noindent
The first term uses the teacher prediction as a hard pseudo-label for the student, whereas the second term uses the student prediction as a hard pseudo-label for the teacher. Thus, both models are updated during joint optimization.

For labeled samples with ground-truth annotation $\mathbf{l}$, both predictions are additionally supervised through

\begin{equation}
\label{eq:loss-sup}
\mathcal{L}_{\mathrm{sup}}
=
\ell_{\mathrm{seg}}
\left(
    \hat{\mathbf{m}}^{(\mathrm{teacher})},
    \mathbf{l}
\right)
+
\ell_{\mathrm{seg}}
\left(
    \hat{\mathbf{m}}^{(\mathrm{student})},
    \mathbf{l}
\right).
\end{equation}

\noindent
The complete co-training objective additionally incorporates the reconstruction and alignment terms introduced by the multi-round extension.

\begin{table*}[!t]
    \centering
    \footnotesize

    \setlength{\tabcolsep}{2.4pt}
    \renewcommand{\arraystretch}{0.95}

    \begin{adjustbox}{max width=\linewidth,center}
    \begin{tabular}{
        @{}
        c
        l
        !{\color{black!15}\vrule width 0.35pt}
        ccccc
        !{\color{black!15}\vrule width 0.35pt}
        ccccc
        !{\color{black!15}\vrule width 0.35pt}
        ccccc
        @{}
    }

        \toprule

        \multicolumn{2}{c}{}
        &
        \multicolumn{5}{>{\columncolor{glascolor}}c}{\textbf{GlaS}}
        &
        \multicolumn{5}{>{\columncolor{ph2color}}c}{\textbf{PH2}}
        &
        \multicolumn{5}{>{\columncolor{hmepscolor}}c}{\textbf{HMEPS}}
        \\

        \cmidrule(lr){3-7}
        \cmidrule(lr){8-12}
        \cmidrule(lr){13-17}

        \multicolumn{2}{c}{}
        &
        \multicolumn{15}{c}{\textbf{Labeled \%}}
        \\[-0.2mm]

        \multicolumn{2}{c}{}
        &
        \cellcolor{glascolor}\textbf{1\%}
        & \cellcolor{glascolor}\textbf{2\%}
        & \cellcolor{glascolor}\textbf{5\%}
        & \cellcolor{glascolor}\textbf{10\%}
        & \cellcolor{glascolor}\textbf{20\%}
        &
        \cellcolor{ph2color}\textbf{1\%}
        & \cellcolor{ph2color}\textbf{2\%}
        & \cellcolor{ph2color}\textbf{5\%}
        & \cellcolor{ph2color}\textbf{10\%}
        & \cellcolor{ph2color}\textbf{20\%}
        &
        \cellcolor{hmepscolor}\textbf{1\%}
        & \cellcolor{hmepscolor}\textbf{2\%}
        & \cellcolor{hmepscolor}\textbf{5\%}
        & \cellcolor{hmepscolor}\textbf{10\%}
        & \cellcolor{hmepscolor}\textbf{20\%}
        \\

        \midrule


        \rowcolor{black!3}
        \multicolumn{17}{c}{\textbf{Dice Coefficient (DC) $\uparrow$}}
        \\
        \cmidrule(lr){1-17}

        \multirow{9}{*}{\rotatebox[origin=c]{90}{\textbf{Method}}}
        &
        \textit{Fully Sup. (100\%)}
        &
        \multicolumn{5}{c}{90.62}
        &
        \multicolumn{5}{c}{92.44}
        &
        \multicolumn{5}{c}{96.98}
        \\[0.3mm]

        \cmidrule(lr){2-17}

        &
        SuperpixSSL
        & 68.51 & 70.10 & 75.18 & 77.82 & 81.33
        & 70.78 & 77.61 & 81.78 & 85.21 & 88.31
        & 87.45 & 88.47 & 91.34 & 92.82 & 93.18
        \\

        &
        RAD-DINO
        & 68.06 & 68.52 & 73.40 & 76.51 & 82.18
        & 77.18 & 79.66 & 85.05 & 88.01 & 91.52
        & 86.69 & 85.81 & 88.72 & 90.10 & 91.17
        \\

        &
        EM
        & 68.92 & 70.23 & 75.34 & 78.08 & 81.20
        & 73.24 & 79.34 & 81.89 & 84.94 & 86.30
        & 90.24 & 91.35 & 92.31 & 92.56 & 93.04
        \\

        &
        CCT
        & 68.97 & 70.05 & 76.33 & \textbf{80.36} & 84.22
        & 73.42 & 80.13 & 82.78 & 87.93 & 89.95
        & 91.09 & 91.48 & 93.25 & 93.45 & 93.30
        \\

        &
        UAMT
        & 69.12 & 69.71 & 75.14 & 79.31 & 83.03
        & 74.72 & 79.76 & 83.75 & 87.37 & 88.95
        & 90.18 & 92.42 & 93.03 & 93.14 & 93.42
        \\

        &
        URPC
        & 68.38 & 68.67 & 74.32 & 78.59 & 82.34
        & 71.23 & 78.88 & 83.41 & 88.06 & 91.73
        & 89.15 & 89.78 & 90.85 & 90.96 & 91.30
        \\

        &
        CPS
        & 69.32 & 70.60 & 76.17 & 80.35 & 83.90
        & 76.07 & 79.64 & 82.86 & 84.33 & 86.49
        & 90.39 & 91.07 & 92.89 & 92.83 & 93.04
        \\

        \rowcolor{ourscolor}
        &
        \textbf{Ours}
        & \textbf{70.03}
        & \textbf{72.22}
        & \textbf{77.30}
        & 80.06
        & \textbf{84.55}
        & \textbf{78.18}
        & \textbf{81.51}
        & \textbf{86.15}
        & \textbf{89.45}
        & \textbf{92.42}
        & \textbf{91.69}
        & \textbf{92.86}
        & \textbf{93.79}
        & \textbf{94.12}
        & \textbf{94.37}
        \\


        \specialrule{0.4pt}{1.0mm}{0pt}
        \specialrule{0.4pt}{0.8pt}{1.0mm}


        \rowcolor{black!3}
        \multicolumn{17}{c}{\textbf{Jaccard Index (JI) $\uparrow$}}
        \\
        \cmidrule(lr){1-17}

        \multirow{9}{*}{\rotatebox[origin=c]{90}{\textbf{Method}}}
        &
        \textit{Fully Sup. (100\%)}
        &
        \multicolumn{5}{c}{82.85}
        &
        \multicolumn{5}{c}{85.96}
        &
        \multicolumn{5}{c}{94.70}
        \\[0.3mm]

        \cmidrule(lr){2-17}

        &
        SuperpixSSL
        & 52.11 & 54.00 & 60.28 & 63.78 & 68.60
        & 54.87 & 63.59 & 69.23 & 74.28 & 79.10
        & 77.80 & 79.53 & 84.16 & 86.65 & 87.23
        \\

        &
        RAD-DINO
        & 51.59 & 52.11 & 57.15 & 61.46 & 68.47
        & 63.19 & 66.22 & 74.05 & 78.02 & 84.38
        & 75.06 & 75.16 & 79.73 & 81.99 & 83.77
        \\

        &
        EM
        & 52.60 & 54.16 & 60.45 & 64.06 & 68.38
        & 57.92 & 65.95 & 69.37 & 73.85 & 75.92
        & 82.25 & 84.14 & 85.78 & 86.20 & 87.00
        \\

        &
        CCT
        & 52.65 & 53.92 & 61.76 & \textbf{67.19} & 72.76
        & 58.06 & 66.90 & 70.66 & 78.46 & 81.74
        & 84.03 & 84.34 & 87.38 & 87.70 & 87.45
        \\

        &
        UAMT
        & 52.83 & 53.55 & 60.19 & 65.72 & 71.00
        & 59.70 & 66.38 & 72.08 & 77.59 & 80.12
        & 82.12 & 86.06 & 87.02 & 87.16 & 87.68
        \\

        &
        URPC
        & 51.96 & 52.31 & 59.17 & 64.78 & 70.12
        & 55.39 & 65.22 & 71.65 & 78.89 & 84.71
        & 80.45 & 81.47 & 83.23 & 83.44 & 84.05
        \\

        &
        CPS
        & 53.05 & 54.59 & 61.54 & 67.16 & 72.27
        & 61.50 & 66.39 & 70.78 & 72.93 & 76.23
        & 82.49 & 83.60 & 86.72 & 86.63 & 86.99
        \\

        \rowcolor{ourscolor}
        &
        \textbf{Ours}
        & \textbf{53.92}
        & \textbf{56.58}
        & \textbf{63.01}
        & 66.69
        & \textbf{73.23}
        & \textbf{64.21}
        & \textbf{68.84}
        & \textbf{75.69}
        & \textbf{80.94}
        & \textbf{85.90}
        & \textbf{84.66}
        & \textbf{86.69}
        & \textbf{88.12}
        & \textbf{88.90}
        & \textbf{89.32}
        \\

        \bottomrule

    \end{tabular}
    \end{adjustbox}

    \caption{\textbf{Quantitative comparison with SOTA methods on GlaS, PH2, and HMEPS.} Mean DC and JI scores (\%) over 10 cross-validation folds across labeling ratios. Best results are shown in bold.}
    \label{tab:sota-comparison}
\end{table*}

\subsection{Multi-round Reconstruction and Segmentation}
\label{sec:multiple_rounds}

The basic co-training formulation in Sec.~\ref{sec:co_training} uses only the mask pathways. To additionally exploit the teacher image pathway during co-training, we generate $R$ stochastic reconstruction--segmentation pairs. The corresponding rounds are independent conditional branches rather than the timesteps of a sequential reverse-diffusion chain.

Starting from the initial teacher prediction $\hat{\mathbf{m}}^{(\mathrm{teacher})}$, each round $r$ independently samples a timestep $t'_r$ and Gaussian noise tensor $\boldsymbol{\nu}^{(r)}_i$. The input image is corrupted as

\begin{equation}
\tilde{\mathbf{i}}^{(r)}_{t'_r}
=
\sqrt{\bar{\alpha}_{t'_r}}\,\mathbf{i}
+
\sqrt{1-\bar{\alpha}_{t'_r}}\,
\boldsymbol{\nu}^{(r)}_i.
\label{eq:round_image_corruption}
\end{equation}

\noindent
The teacher image pathway predicts the corresponding noise,

\begin{equation}
\hat{\boldsymbol{\nu}}^{(r)}_i
=
f_i^{(\mathrm{teacher})}
\left(
    \tilde{\mathbf{i}}^{(r)}_{t'_r},
    \hat{\mathbf{m}}^{(\mathrm{teacher})},
    t'_r
\right),
\label{eq:round_noise_prediction}
\end{equation}

\noindent
and produces the clean-image estimate

\begin{equation}
\hat{\mathbf{i}}_r
=
g_i^{(\mathrm{teacher})}
\left(
    \tilde{\mathbf{i}}^{(r)}_{t'_r},
    \hat{\mathbf{m}}^{(\mathrm{teacher})},
    t'_r
\right).
\label{eq:round_image_reconstruction}
\end{equation}

\noindent
A second independent timestep $t''_r$ and Gaussian tensor $\boldsymbol{\nu}^{(r)}_m$ are used to construct a corrupted empty-mask initialization:

\begin{equation}
\tilde{\mathbf{m}}^{(\emptyset,r)}_{t''_r}
=
\sqrt{\bar{\alpha}_{t''_r}}\,
\mathbf{m}^{(\emptyset)}
+
\sqrt{1-\bar{\alpha}_{t''_r}}\,
\boldsymbol{\nu}^{(r)}_m.
\label{eq:round_mask_corruption}
\end{equation}

\noindent
The reconstructed image is then processed by the teacher mask pathway to obtain an additional segmentation prediction:

\begin{equation}
\hat{\mathbf{m}}_r
=
f_m^{(\mathrm{teacher})}
\left(
    \hat{\mathbf{i}}_r,
    \tilde{\mathbf{m}}^{(\emptyset,r)}_{t''_r},
    t''_r
\right).
\label{eq:round_mask_prediction}
\end{equation}

The additional mask prediction is optimized through the alignment objective

\begin{equation}
\mathcal{L}_r^{(\mathrm{align})}
=
\ell_{\mathrm{seg}}
\left(
    \hat{\mathbf{m}}_r,
    \hat{\mathbf{l}}
\right),
\label{eq:round_alignment_loss}
\end{equation}

\noindent
where

\begin{equation}
\hat{\mathbf{l}}
=
\begin{cases}
\mathbf{l},
& \text{labeled},
\\
\operatorname*{arg\,max}_{c}
\hat{\mathbf{m}}^{(\mathrm{student})},
& \text{unlabeled}.
\end{cases}
\label{eq:alignment_target}
\end{equation}

\noindent
Consistently with the image-pathway objective used during pretraining, the reconstruction loss for each round is defined in noise-prediction space:

\begin{equation}
\mathcal{L}_r^{(\mathrm{rec})}
=
\operatorname{MSE}
\left(
    \boldsymbol{\nu}^{(r)}_i,
    \hat{\boldsymbol{\nu}}^{(r)}_i
\right).
\label{eq:round_reconstruction_loss}
\end{equation}

The $R$ rounds use independently sampled noise tensors and timesteps and can therefore be computed in parallel. For notational simplicity, each loss term below denotes an average over the subset of samples to which it applies. For a minibatch containing labeled and unlabeled samples, the complete training objective is

\begin{equation}
\begin{aligned}
\mathcal{L}
={}&
\mathcal{L}_{\mathrm{sup}}
+
\lambda_{\mathrm{diff}}
\frac{1}{R}
\sum_{r=1}^{R}
\left(
\mathcal{L}_{r,L}^{(\mathrm{align})}
+
\mathcal{L}_{r,L}^{(\mathrm{rec})}
\right)
\\
&+
\lambda_{\mathrm{u}}
\left[
\mathcal{L}_{\mathrm{CPS}}
+
\lambda_{\mathrm{diff}}
\frac{1}{R}
\sum_{r=1}^{R}
\left(
\mathcal{L}_{r,U}^{(\mathrm{align})}
+
\mathcal{L}_{r,U}^{(\mathrm{rec})}
\right)
\right].
\end{aligned}
\label{eq:complete_objective}
\end{equation}

\noindent
Here, the subscripts $L$ and $U$ denote loss terms evaluated on labeled and unlabeled samples, respectively. As defined in Eq.~\ref{eq:alignment_target}, alignment uses the ground truth for labeled samples and the hard student pseudo-label for unlabeled ones. The weights $\lambda_{\mathrm{u}}$ and $\lambda_{\mathrm{diff}}$ are specified in the suppl. material.

\begin{figure*}[!t]
    \centering
    \small

    \setlength{\tabcolsep}{2.5pt}
    \renewcommand{\arraystretch}{1.0}

    \begin{minipage}[t]{0.45\textwidth}
        \centering

        \begin{tcolorbox}[
            enhanced,
            width=\linewidth,
            colback=white,
            colframe={rgb,255:red,92;green,145;blue,190},
            boxrule=1.1pt,
            arc=2.5mm,
            left=1.2mm,
            right=1.2mm,
            top=1.8mm,
            bottom=0.6mm,
            boxsep=0pt,
            before skip=0pt,
            after skip=0pt,
            title=\textbf{GlaS},
            attach boxed title to top center={
                yshift=-2.0mm
            },
            boxed title style={
                colback=white,
                colframe=white,
                boxrule=0pt,
                left=2.5mm,
                right=2.5mm,
                top=0.3mm,
                bottom=0.3mm
            },
            coltitle={rgb,255:red,92;green,145;blue,190},
            fonttitle=\small\bfseries
        ]

        \centering
        \begin{tabular}{ccc}
            {\footnotesize\textbf{Sample}} &
            {\footnotesize\textbf{Prediction}} &
            {\footnotesize\textbf{Target}}
            \\[-1.9mm]

            \textcolor[RGB]{92,145,190}{\rule{1.65cm}{0.35pt}} &
            \textcolor[RGB]{92,145,190}{\rule{1.65cm}{0.35pt}} &
            \textcolor[RGB]{92,145,190}{\rule{1.65cm}{0.35pt}}
            \\[0mm]

            \includegraphics[
                width=2.35cm,
                height=1.55cm
            ]{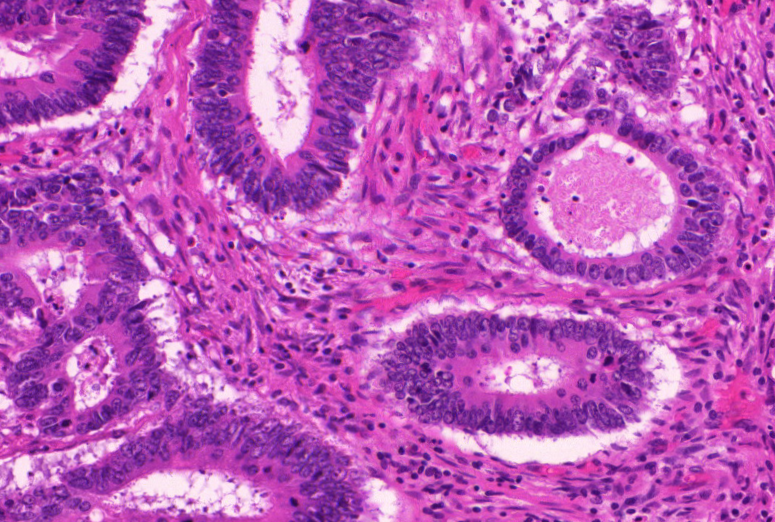}
            &
            \includegraphics[
                width=2.35cm,
                height=1.55cm
            ]{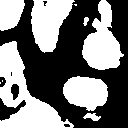}
            &
            \includegraphics[
                width=2.35cm,
                height=1.55cm
            ]{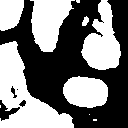}
        \end{tabular}

        \end{tcolorbox}
    \end{minipage}%
    \hspace{1.0mm}%
    \begin{minipage}[t]{0.45\textwidth}
        \centering

        \begin{tcolorbox}[
            enhanced,
            width=\linewidth,
            colback=white,
            colframe={rgb,255:red,158;green,120;blue,174},
            boxrule=1.1pt,
            arc=2.5mm,
            left=1.2mm,
            right=1.2mm,
            top=1.8mm,
            bottom=0.6mm,
            boxsep=0pt,
            before skip=0pt,
            after skip=0pt,
            title=\textbf{PH2},
            attach boxed title to top center={
                yshift=-2.0mm
            },
            boxed title style={
                colback=white,
                colframe=white,
                boxrule=0pt,
                left=2.5mm,
                right=2.5mm,
                top=0.3mm,
                bottom=0.3mm
            },
            coltitle={rgb,255:red,158;green,120;blue,174},
            fonttitle=\small\bfseries
        ]

        \centering
        \begin{tabular}{ccc}
            {\footnotesize\textbf{Sample}} &
            {\footnotesize\textbf{Prediction}} &
            {\footnotesize\textbf{Target}}
            \\[-1.9mm]

            \textcolor[RGB]{158,120,174}{\rule{1.65cm}{0.35pt}} &
            \textcolor[RGB]{158,120,174}{\rule{1.65cm}{0.35pt}} &
            \textcolor[RGB]{158,120,174}{\rule{1.65cm}{0.35pt}}
            \\[0mm]

            \includegraphics[
                width=2.35cm,
                height=1.55cm
            ]{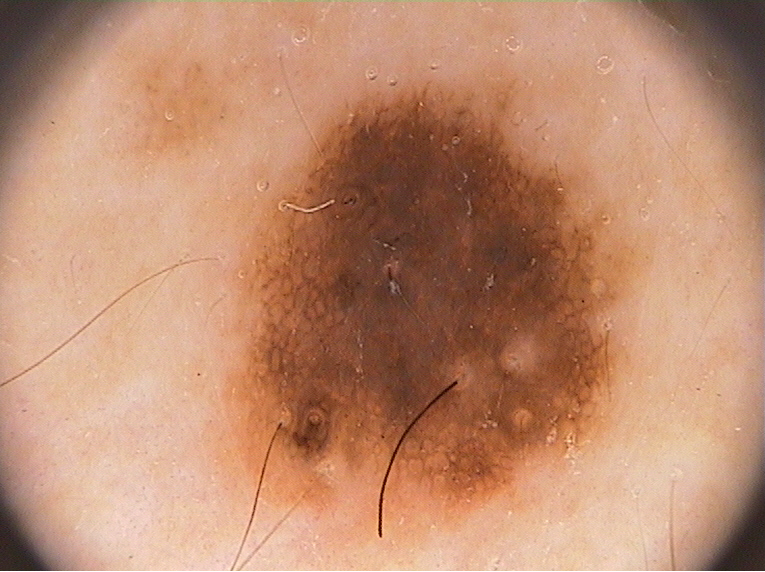}
            &
            \includegraphics[
                width=2.35cm,
                height=1.55cm
            ]{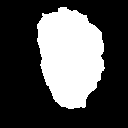}
            &
            \includegraphics[
                width=2.35cm,
                height=1.55cm
            ]{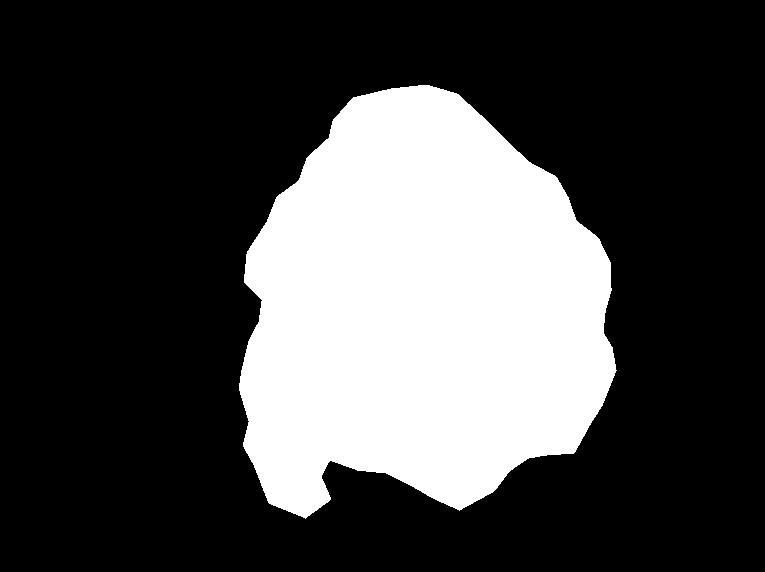}
        \end{tabular}

        \end{tcolorbox}
    \end{minipage}

    \par\vspace{0.4mm}

    \begin{minipage}[t]{0.45\textwidth}
        \centering

        \begin{tcolorbox}[
            enhanced,
            width=\linewidth,
            colback=white,
            colframe={rgb,255:red,96;green,155;blue,122},
            boxrule=1.1pt,
            arc=2.5mm,
            left=1.2mm,
            right=1.2mm,
            top=1.8mm,
            bottom=0.6mm,
            boxsep=0pt,
            before skip=0pt,
            after skip=0pt,
            title=\textbf{HMEPS},
            attach boxed title to top center={
                yshift=-2.0mm
            },
            boxed title style={
                colback=white,
                colframe=white,
                boxrule=0pt,
                left=2.5mm,
                right=2.5mm,
                top=0.3mm,
                bottom=0.3mm
            },
            coltitle={rgb,255:red,96;green,155;blue,122},
            fonttitle=\small\bfseries
        ]

        \centering
        \begin{tabular}{ccc}
            {\footnotesize\textbf{Sample}} &
            {\footnotesize\textbf{Prediction}} &
            {\footnotesize\textbf{Target}}
            \\[-1.9mm]

            \textcolor[RGB]{96,155,122}{\rule{1.65cm}{0.35pt}} &
            \textcolor[RGB]{96,155,122}{\rule{1.65cm}{0.35pt}} &
            \textcolor[RGB]{96,155,122}{\rule{1.65cm}{0.35pt}}
            \\[0mm]

            \includegraphics[
                width=2.35cm,
                height=1.55cm
            ]{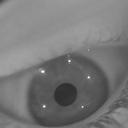}
            &
            \includegraphics[
                width=2.35cm,
                height=1.55cm
            ]{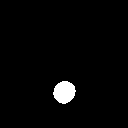}
            &
            \includegraphics[
                width=2.35cm,
                height=1.55cm
            ]{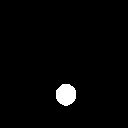}
        \end{tabular}

        \end{tcolorbox}
    \end{minipage}%
    \hspace{1.0mm}%
    \begin{minipage}[t]{0.45\textwidth}
        \centering

        \begin{tcolorbox}[
            enhanced,
            width=\linewidth,
            colback=white,
            colframe={rgb,255:red,190;green,145;blue,86},
            boxrule=1.1pt,
            arc=2.5mm,
            left=1.2mm,
            right=1.2mm,
            top=1.8mm,
            bottom=0.6mm,
            boxsep=0pt,
            before skip=0pt,
            after skip=0pt,
            title=\textbf{LA},
            attach boxed title to top center={
                yshift=-2.0mm
            },
            boxed title style={
                colback=white,
                colframe=white,
                boxrule=0pt,
                left=2.5mm,
                right=2.5mm,
                top=0.3mm,
                bottom=0.3mm
            },
            coltitle={rgb,255:red,190;green,145;blue,86},
            fonttitle=\small\bfseries
        ]

        \centering
        \begin{tabular}{ccc}
            {\footnotesize\textbf{Sample}} &
            {\footnotesize\textbf{Prediction}} &
            {\footnotesize\textbf{Target}}
            \\[-1.9mm]

            \textcolor[RGB]{190,145,86}{\rule{1.65cm}{0.35pt}} &
            \textcolor[RGB]{190,145,86}{\rule{1.65cm}{0.35pt}} &
            \textcolor[RGB]{190,145,86}{\rule{1.65cm}{0.35pt}}
            \\[0mm]

            \includegraphics[
                width=2.35cm,
                height=1.55cm
            ]{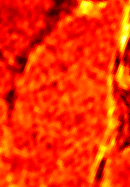}
            &
            \includegraphics[
                width=2.35cm,
                height=1.55cm
            ]{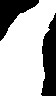}
            &
            \includegraphics[
                width=2.35cm,
                height=1.55cm
            ]{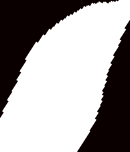}
        \end{tabular}

        \end{tcolorbox}
    \end{minipage}

    \caption{\textbf{Qualitative segmentation results using 20\% of the annotations.}
    For each dataset, we show the input image, final prediction, and ground-truth target. For LA, a representative 2D slice from a predicted 3D volume is shown.}
    \label{fig:qualitative}
\end{figure*}

\section{Experiments}
\label{sec:exp}

\subsection{Datasets and Evaluation Metrics}
\label{sec:sec:datasets}

We evaluate the proposed method on three publicly available datasets for 2D biomedical image segmentation, covering histological, dermoscopic, and infrared eye imaging. We adopt a 10-fold cross-validation protocol, with further details provided in the supplementary material.

\noindent \textbf{GlaS~\cite{SIRINUKUNWATTANA2017489}.}
The gland segmentation in colon histology images challenge (GlaS) dataset consists of 165 Hematoxylin and Eosin (H\&E)-stained colorectal histology images with a resolution of 775$\times$522 pixels. 

\noindent \textbf{PH2~\cite{6610779}.}
The PH2 dataset contains 200 dermoscopic images of melanocytic lesions.
The images are 8-bit RGB images with a resolution of 768$\times$560 pixels and are provided with lesion segmentation masks.

\noindent \textbf{HMEPS~\cite{raffaele_mazziotti_2021_4488164}.}
The human and mouse eyes for pupil semantic segmentation (HMEPS) dataset comprises 11,897 grayscale images of human (4,285) and mouse (7,612) eyes acquired under infrared (IR, 850 nm) illumination. Pupil regions are manually annotated using polygonal masks.

For evaluation, we report the Dice coefficient (DC) and Jaccard index (JI), two commonly used metrics for biomedical image segmentation~\cite{LUO2022102517,Luo_2021,7785132,9706678}. 

\subsection{Comparison with SOTA}
\label{sec:sec:results-comparison-sota}

We quantitatively compare our approach with representative semi-supervised segmentation methods covering entropy minimization, consistency regularization, and pseudo-labeling strategies: EM~\cite{DBLP:conf/cvpr/VuJBCP19}, CCT~\cite{9157032}, UAMT~\cite{10.1007/978-3-030-32245-8_67}, CPS~\cite{9577639}, and URPC~\cite{LUO2022102517}. Since the original works adopt different experimental protocols, we reimplement all methods within a common evaluation framework using the same U-Net backbone while retaining the method-specific auxiliary components required by each approach.

\begin{table}[!t]
    \centering
    \scriptsize

    \setlength{\tabcolsep}{2.2pt}
    \renewcommand{\arraystretch}{0.98}

    \caption{\textbf{Quantitative comparison on the volumetric Left Atrium (LA) dataset.}
    Results are reported as mean DC and JI scores (\%) over five cross-validation folds.
    Best result in bold.}
    \label{tab:atrial}

    \resizebox{\columnwidth}{!}{%
    \begin{tabular}{@{}lccccc@{}}
        \toprule

        &
        \multicolumn{5}{c}{\textbf{Labeled \%}}
        \\[0.6mm]

        &
        \textbf{1\%}
        &
        \textbf{2\%}
        &
        \textbf{5\%}
        &
        \textbf{10\%}
        &
        \textbf{20\%}
        \\[-0.8mm]

        \cmidrule(lr){2-6}

        \raisebox{-0.8ex}[0pt][0pt]{\textbf{Method}}
        &
        \textbf{DC$\uparrow$/JI$\uparrow$}
        &
        \textbf{DC$\uparrow$/JI$\uparrow$}
        &
        \textbf{DC$\uparrow$/JI$\uparrow$}
        &
        \textbf{DC$\uparrow$/JI$\uparrow$}
        &
        \textbf{DC$\uparrow$/JI$\uparrow$}
        \\

        \midrule

        EM
        & 64.00/47.09
        & 73.53/58.36
        & 80.80/67.84
        & 85.33/74.41
        & 89.51/81.01 \\

        CCT
        & 61.12/44.27
        & 75.49/60.63
        & 84.70/73.47
        & 88.11/78.75
        & 90.74/83.05 \\

        UAMT
        & 60.42/43.39
        & 74.80/59.93
        & 84.22/72.80
        & 88.95/80.09
        & \textbf{90.91/83.34} \\

        URPC
        & 68.48/52.14
        & 77.53/63.37
        & 83.26/73.09
        & 86.84/76.75
        & 89.12/80.38 \\

        CPS
        & 71.69/56.57
        & \textbf{83.65/71.91}
        & 87.31/77.48
        & 89.57/81.12
        & 90.41/82.51 \\

        DTC
        & 62.63/45.66
        & 76.87/62.51
        & 83.28/71.43
        & 86.43/76.13
        & 89.46/80.93 \\

        \cellcolor[RGB]{246,246,246}\textbf{Ours}
        & \cellcolor[RGB]{246,246,246}\textbf{73.58/58.33}
        & \cellcolor[RGB]{246,246,246}82.78/70.63
        & \cellcolor[RGB]{246,246,246}\textbf{87.95/78.50}
        & \cellcolor[RGB]{246,246,246}\textbf{89.70/81.32}
        & \cellcolor[RGB]{246,246,246}90.61/82.83 \\

        \bottomrule
    \end{tabular}%
    }
\end{table}

We additionally consider two-stage representation-learning baselines based on self-supervised pretraining~\cite{10.1007/978-3-030-58526-6_45,perez2025}. These approaches first learn visual representations from unlabeled images and subsequently adapt them to the segmentation task using the available labeled samples. We specifically evaluate SuperpixSSL~\cite{10.1007/978-3-030-58526-6_45}, which learns representations using superpixel-derived pseudo-labels, and RAD-DINO~\cite{perez2025}, which employs a transformer-based encoder~\cite{dosovitskiy2020,oquab2023} and is therefore the only baseline in our comparison that does not use a U-Net-based backbone. For reference, we also include a fully supervised U-Net trained using all available annotations. 

In the semi-supervised experiments, we retain $1\%$, $2\%$, $5\%$, $10\%$, or $20\%$ of the training images as labeled samples and treat the remaining training images as unlabeled. Results are averaged over 10 cross-validation folds.
Further details on training configurations and baseline implementations are provided in the supplementary material. Quantitative results are reported in Table~\ref{tab:sota-comparison}, while qualitative examples are shown in Fig.~\ref{fig:qualitative}.

\begin{figure*}[!t]
    \centering
    \small

    \makebox[\linewidth][c]{%
        \begin{minipage}[c]{0.028\linewidth}
            \mbox{}
        \end{minipage}%
        \hspace{0.8mm}%
        \begin{minipage}[c]{0.265\linewidth}
            \centering
            \textbf{GlaS}
        \end{minipage}%
        \hspace{1.5mm}%
        \begin{minipage}[c]{0.265\linewidth}
            \centering
            \textbf{PH2}
        \end{minipage}%
        \hspace{1.5mm}%
        \begin{minipage}[c]{0.265\linewidth}
            \centering
            \textbf{HMEPS}
        \end{minipage}%
    }

    \vspace{0.5mm}

    \makebox[\linewidth][c]{%
        \begin{minipage}[c]{0.028\linewidth}
            \makebox[\linewidth][r]{%
                \rotatebox[origin=c]{90}{%
                    \footnotesize\textbf{Teacher pretraining}%
                }%
            }
        \end{minipage}%
        \hspace{0.8mm}%
        \begin{minipage}[c]{0.265\linewidth}
            \centering
            \includegraphics[width=\linewidth]{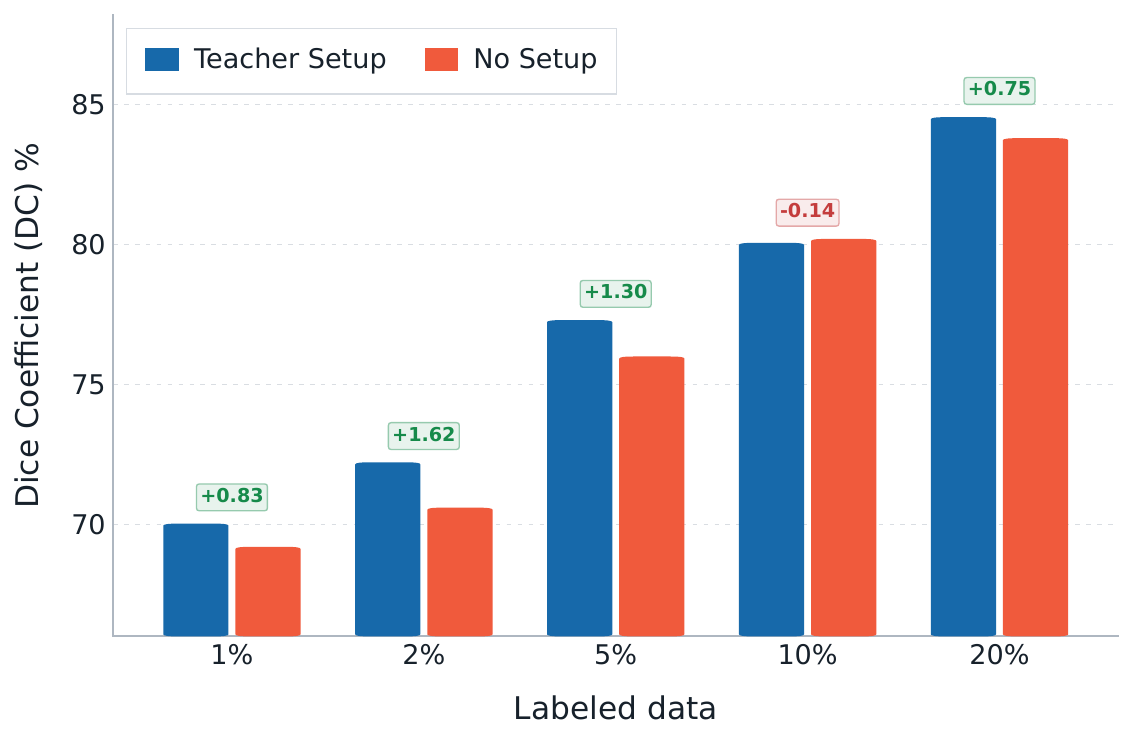}
        \end{minipage}%
        \hspace{1.5mm}%
        \begin{minipage}[c]{0.265\linewidth}
            \centering
            \includegraphics[width=\linewidth]{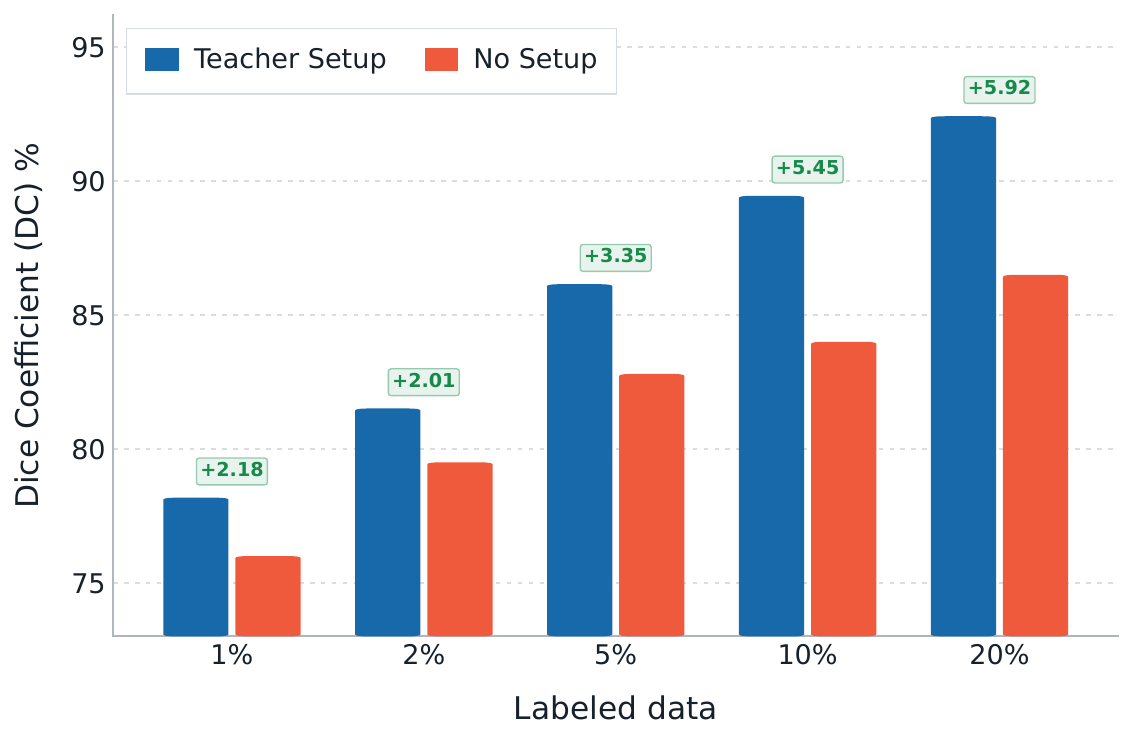}
        \end{minipage}%
        \hspace{1.5mm}%
        \begin{minipage}[c]{0.265\linewidth}
            \centering
            \includegraphics[width=\linewidth]{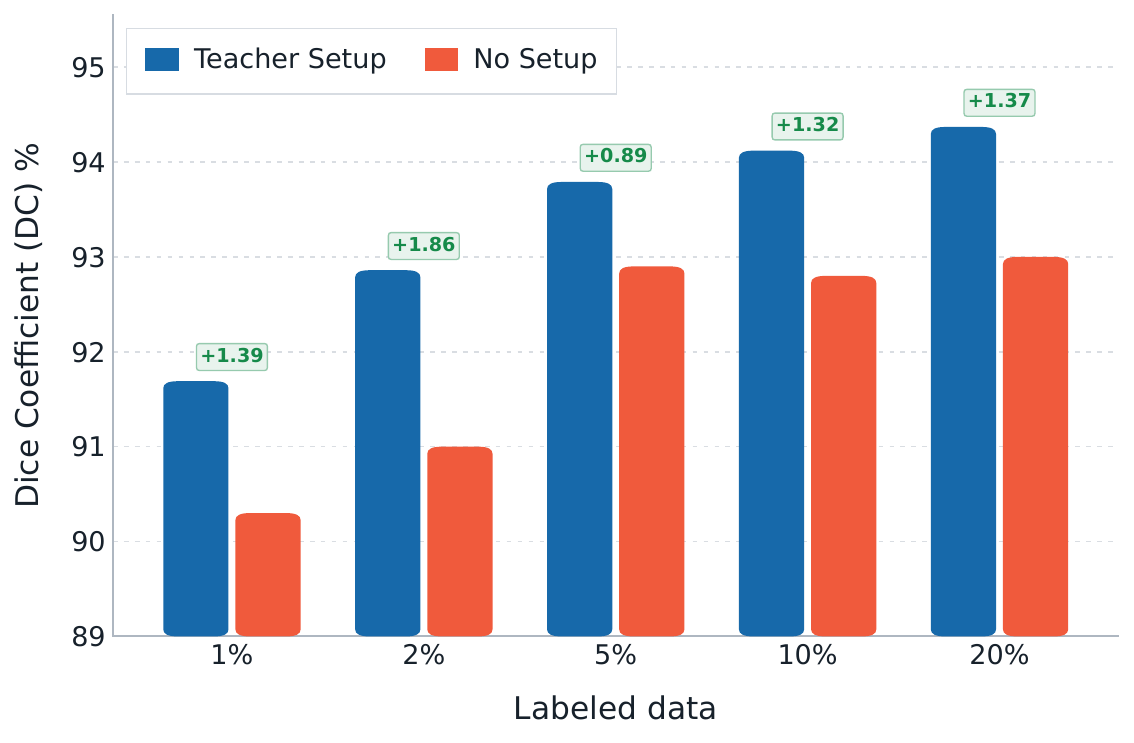}
        \end{minipage}%
    }

    \vspace{0.7mm}

    \makebox[\linewidth][c]{%
        \begin{minipage}[c]{0.028\linewidth}
            \makebox[\linewidth][r]{%
                \rotatebox[origin=c]{90}{%
                    \footnotesize\textbf{Diffusion rounds}%
                }%
            }
        \end{minipage}%
        \hspace{0.8mm}%
        \begin{minipage}[c]{0.265\linewidth}
            \centering
            \includegraphics[width=\linewidth]{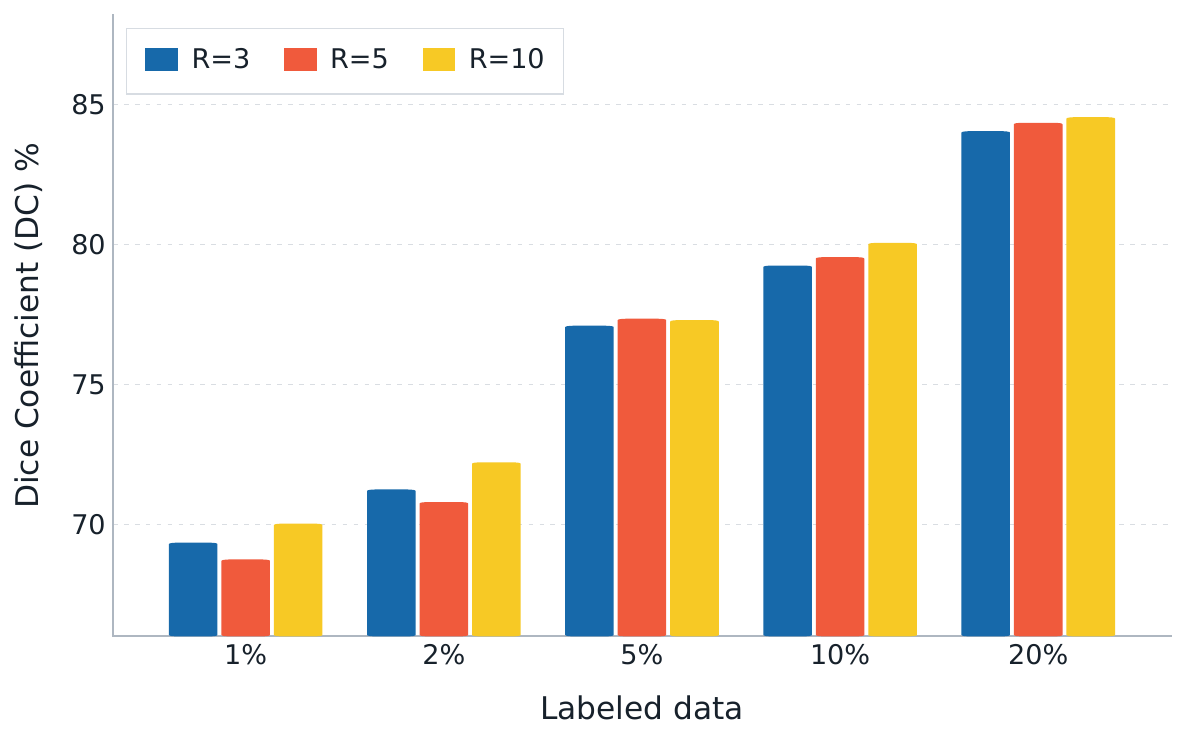}
        \end{minipage}%
        \hspace{1.5mm}%
        \begin{minipage}[c]{0.265\linewidth}
            \centering
            \includegraphics[width=\linewidth]{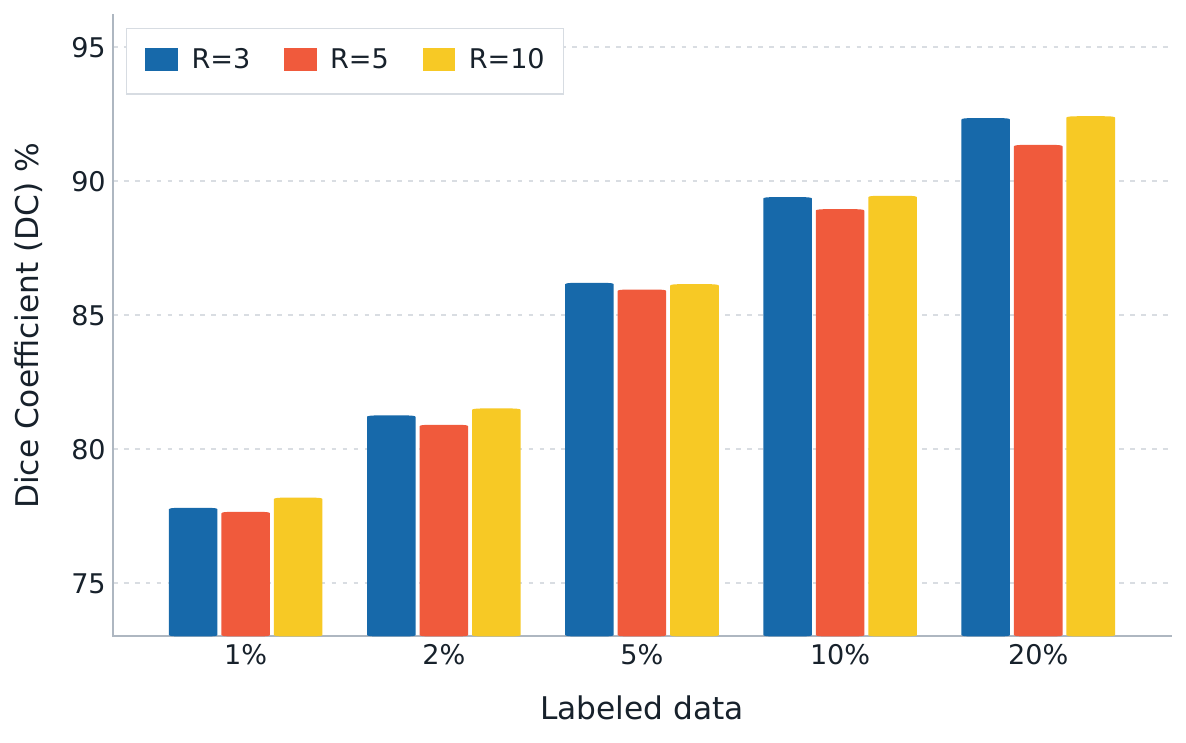}
        \end{minipage}%
        \hspace{1.5mm}%
        \begin{minipage}[c]{0.265\linewidth}
            \centering
            \includegraphics[width=\linewidth]{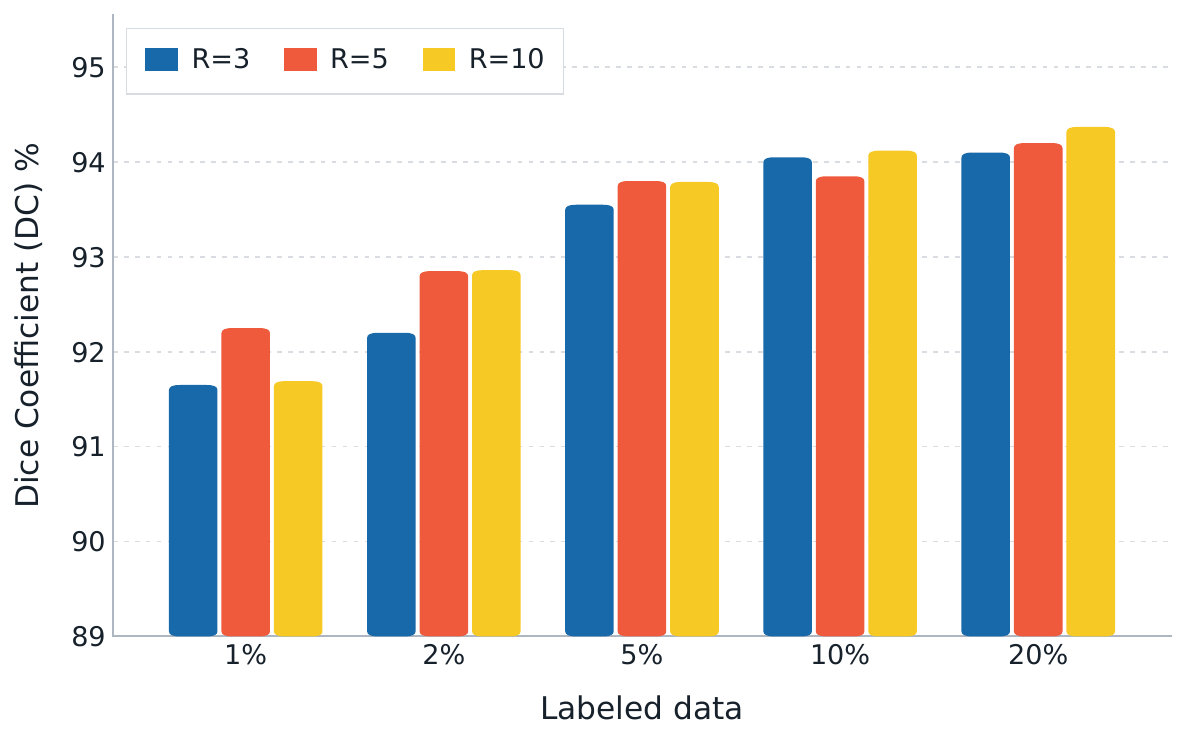}
        \end{minipage}%
    }

    \caption{\textbf{Ablation studies on teacher pretraining and the number of diffusion rounds $\mathbf{R}$.}
    Top: DC with and without reconstruction-based teacher pretraining; values above the bars indicate the performance difference.
    Bottom: DC for $R \in \{3,5,10\}$.
    Results are shown on GlaS, PH2, and HMEPS across labeling ratios.}
    \label{fig:ablation}
\end{figure*}

\paragraph{GlaS.}
On the GlaS dataset, the proposed method achieves the best performance in terms of both DC and JI for four of the five labeling ratios considered. The improvements are particularly evident in the most label-scarce settings, from $1\%$ to $5\%$, indicating that the proposed approach effectively exploits unlabeled images when only a limited number of annotations is available. At the $10\%$ labeling ratio, CCT and CPS obtain slightly higher scores, although our method remains competitive, with differences of at most 0.3 percentage points in DC and 0.5 percentage points in JI. At $20\%$, our approach again achieves the best results.

\paragraph{PH2.}
On PH2, the proposed method outperforms all competing semi-supervised approaches in terms of both DC and JI across all labeling ratios. The improvements are observed consistently from the most label-scarce setting to the $20\%$ configuration. Notably, when using only $20\%$ of the available annotations, our method reaches $92.42\%$ DC and $85.90\%$ JI, closely matching the fully supervised baseline.

\paragraph{HMEPS.}
On HMEPS, the proposed method consistently outperforms all competing semi-supervised approaches in terms of both DC and JI across all labeling ratios. The advantage is maintained from the $1\%$ setting to the $20\%$ setting, where our method reaches $94.37\%$ DC and $89.32\%$ JI. 

\subsection{Ablation Studies}
\label{sec:sec:ablation}

We ablates the main components and configurations of our method on all three 2D benchmarks across all labeling ratios using DC.

\paragraph{Effect of teacher pretraining.}
We first evaluate the contribution of the unsupervised reconstruction-based teacher pretraining stage by comparing the complete method with a variant in which the teacher is initialized without the proposed pretraining. As shown in Fig.~\ref{fig:ablation} (top), pretraining improves DC performance in nearly all evaluated configurations, with particularly consistent gains on PH2 and HMEPS.

\paragraph{Number of diffusion rounds $\mathbf{R}$.}
We analyze the effect of $R$, which controls the number of stochastic reconstruction--segmentation pairs introduced in Sec.~\ref{sec:multiple_rounds}. As shown in Fig.~\ref{fig:ablation} (bottom), performance is relatively stable across the evaluated values of $R$, with no consistent monotonic trend across datasets and labeling ratios. Among the tested configurations, $R=10$ provides the strongest overall performance and is therefore used in the main experiments. However, the differences between $R=5$ and $R=10$ are generally modest, suggesting that $R=5$ can provide a lower-cost alternative when computational efficiency is prioritized.

\subsection{Experiments with Volumetric Images}
\label{sec:sec:3d-exp}

To further assess the applicability of the proposed framework beyond 2D segmentation, we conduct experiments on the volumetric Left Atrium (LA) dataset~\cite{XIONG2021101832}. The dataset contains 100 3D MRI volumes from the 2018 Atrial Segmentation Challenge. We evaluate the methods using 5-fold cross-validation, with further details provided in the supplementary material. For this volumetric setting, we compare against EM, CCT, UAMT, URPC, CPS, and DTC~\cite{Luo_2021}. Quantitative results are reported in Table~\ref{tab:atrial}, while qualitative examples are provided in Fig.~\ref{fig:qualitative}.
The results are consistent with the trends observed on the 2D benchmarks. Our method achieves the best DC and JI scores at the 1\%, 5\%, and 10\% labeling ratios. At 2\%, CPS obtains the highest scores, while our method remains within 0.87 percentage points in DC. At 20\%, UAMT performs best, with our method within 0.30 points in DC.

\section{Conclusion}
\label{sec:conclusions}

In this work, we introduced a semi-supervised framework for biomedical image segmentation that combines teacher--student co-training with the forward diffusion process, timestep conditioning, and conditional denoising objectives used in DDPMs. The teacher is first pretrained through an unsupervised reconstruction task in which an intermediate segmentation prediction conditions image denoising, encouraging the predicted mask to retain information useful for recovering the input image. The pretrained teacher and the student are then jointly optimized using supervised segmentation on labeled samples and cross pseudo-supervision on unlabeled data. A multi-round extension further exploits the teacher image pathway during co-training by generating stochastic reconstruction--segmentation pairs that provide additional reconstruction and alignment signals.

We evaluated the proposed framework on three public 2D biomedical segmentation datasets, GlaS, PH2, and HMEPS, and on the volumetric LA benchmark for left atrial segmentation. Across datasets and labeling ratios, the proposed method achieved competitive or superior performance compared with representative semi-supervised approaches. The ablation studies further showed the benefit of reconstruction-based teacher pretraining and examined the effect of the number of reconstruction--segmentation rounds.
Overall, the results support the use of diffusion-based corruption and denoising objectives as complementary training signals within teacher--student semi-supervised segmentation. 

{
    \small
    \bibliographystyle{ieeenat_fullname}
    \bibliography{main}

@String(CVPR= {IEEE Conf. Comput. Vis. Pattern Recog.})

@String(ICCV= {Int. Conf. Comput. Vis.})

@String(ECCV= {Eur. Conf. Comput. Vis.})

@String(NIPS= {Adv. Neural Inform. Process. Syst.})

@String(AAAI = {AAAI})

@String(CVPR  = {CVPR})

@String(ICCV  = {ICCV})

@String(ECCV  = {ECCV})

@String(NIPS  = {NeurIPS})

@InProceedings{10.1007/978-3-319-24574-4_28,
    author="Ronneberger, Olaf
    and Fischer, Philipp
    and Brox, Thomas",
    title="U-Net: Convolutional Networks for Biomedical Image Segmentation",
    booktitle="Medical Image Computing and Computer-Assisted Intervention -- MICCAI 2015",
    year="2015",
    publisher="Springer International Publishing",
    address="Cham",
    pages="234--241",
    doi="10.1007/978-3-319-24574-4\_28",
    isbn="978-3-319-24574-4"
}

@article{DBLP:journals/corr/ChenPSA17,
  author       = {Liang{-}Chieh Chen and
                  George Papandreou and
                  Florian Schroff and
                  Hartwig Adam},
  title        = {Rethinking Atrous Convolution for Semantic Image Segmentation},
  journal      = {CoRR},
  volume       = {abs/1706.05587},
  year         = {2017},
  eprinttype    = {arXiv},
  eprint       = {1706.05587},
}

@inproceedings{7298965,
  author       = {Jonathan Long and
                  Evan Shelhamer and
                  Trevor Darrell},
  title        = {Fully convolutional networks for semantic segmentation},
  booktitle    = {{IEEE} Conference on Computer Vision and Pattern Recognition, {CVPR}
                  2015, Boston, MA, USA, June 7-12, 2015},
  pages        = {3431--3440},
  publisher    = {{IEEE} Computer Society},
  year         = {2015},
  doi          = {10.1109/CVPR.2015.7298965},
}

@INPROCEEDINGS {7785132,
    author = {F. Milletari and N. Navab and S. Ahmadi},
    booktitle = {2016 Fourth International Conference on 3D Vision (3DV)},
    title = {V-Net: Fully Convolutional Neural Networks for Volumetric Medical Image Segmentation},
    year = {2016},
    volume = {},
    issn = {},
    pages = {565-571},
    doi = {10.1109/3DV.2016.79},
    publisher = {IEEE Computer Society},
    address = {Los Alamitos, CA, USA},
    month = {oct}
}

@InProceedings{10.1007/978-3-030-87199-4_16,
    author="Xie, Yutong
    and Zhang, Jianpeng
    and Shen, Chunhua
    and Xia, Yong",
    title="CoTr: Efficiently Bridging CNN and Transformer for 3D Medical Image Segmentation",
    booktitle="Medical Image Computing and Computer Assisted Intervention -- MICCAI 2021",
    year="2021",
    publisher="Springer International Publishing",
    address="Cham",
    pages="171--180",
    isbn="978-3-030-87199-4",
    doi={10.1007/978-3-030-87199-4\_16}
}

@INPROCEEDINGS{9706678,
  author={Hatamizadeh, Ali and Tang, Yucheng and Nath, Vishwesh and Yang, Dong and Myronenko, Andriy and Landman, Bennett and Roth, Holger R. and Xu, Daguang},
  booktitle={2022 IEEE/CVF Winter Conference on Applications of Computer Vision (WACV)}, 
  title={UNETR: Transformers for 3D Medical Image Segmentation}, 
  year={2022},
  volume={},
  number={},
  pages={1748-1758},
  doi={10.1109/WACV51458.2022.00181}
}

@InProceedings{10.1007/978-3-031-25066-89,
    author="Cao, Hu
    and Wang, Yueyue
    and Chen, Joy
    and Jiang, Dongsheng
    and Zhang, Xiaopeng
    and Tian, Qi
    and Wang, Manning",
    title="Swin-Unet: Unet-Like Pure Transformer for Medical Image Segmentation",
    booktitle="Computer Vision -- ECCV 2022 Workshops",
    year="2023",
    publisher="Springer Nature Switzerland",
    address="Cham",
    pages="205--218",
    isbn="978-3-031-25066-8",
    doi={10.1007/978-3-031-25066-8\_9}
}

@ARTICLE{9625988,
  author={Valanarasu, Jeya Maria Jose and Sindagi, Vishwanath A. and Hacihaliloglu, Ilker and Patel, Vishal M.},
  journal={IEEE Transactions on Medical Imaging}, 
  title={KiU-Net: Overcomplete Convolutional Architectures for Biomedical Image and Volumetric Segmentation}, 
  year={2022},
  volume={41},
  number={4},
  pages={965-976},
  doi={10.1109/TMI.2021.3130469}
}

@article{Isensee_2020, 
    title={nnU-Net: a self-configuring method for deep learning-based biomedical image segmentation}, 
    volume={18}, 
    ISSN={1548-7105}, 
    DOI={10.1038/s41592-020-01008-z}, 
    number={2}, 
    journal={Nature Methods}, 
    publisher={Springer Science and Business Media LLC}, 
    author={Isensee, Fabian and Jaeger, Paul F. and Kohl, Simon A. A. and Petersen, Jens and Maier-Hein, Klaus H.}, 
    year={2020}, 
    month=dec, 
    pages={203–211} 
}

@InProceedings{10.1007/978-3-319-46723-8_49,
    author="{\c{C}}i{\c{c}}ek, {\"O}zg{\"u}n
    and Abdulkadir, Ahmed
    and Lienkamp, Soeren S.
    and Brox, Thomas
    and Ronneberger, Olaf",
    title="3D U-Net: Learning Dense Volumetric Segmentation from Sparse Annotation",
    booktitle="Medical Image Computing and Computer-Assisted Intervention -- MICCAI 2016",
    year="2016",
    publisher="Springer International Publishing",
    address="Cham",
    pages="424--432",
    isbn="978-3-319-46723-8",
    doi = {10.1007/978-3-319-46723-8\_49},
}

@InProceedings{10.1007/978-3-030-87193-2_4,
    author="Valanarasu, Jeya Maria Jose
    and Oza, Poojan
    and Hacihaliloglu, Ilker
    and Patel, Vishal M.",
    title="Medical Transformer: Gated Axial-Attention for Medical Image Segmentation",
    booktitle="Medical Image Computing and Computer Assisted Intervention -- MICCAI 2021",
    year="2021",
    publisher="Springer International Publishing",
    address="Cham",
    pages="36--46",
    isbn="978-3-030-87193-2",
    doi={10.1007/978-3-030-87193-2\_4}
}

@article{10.1016/j.patcog.2022.108673,
    author = {Basak, Hritam and Kundu, Rohit and Sarkar, Ram},
    title = {MFSNet: A multi focus segmentation network for skin lesion segmentation},
    year = {2022},
    issue_date = {Aug 2022},
    publisher = {Elsevier Science Inc.},
    address = {USA},
    volume = {128},
    number = {C},
    issn = {0031-3203},
    doi = {10.1016/j.patcog.2022.108673},
    journal = {Pattern Recogn.},
    month = {aug},
    numpages = {12},
}

@ARTICLE{10332179,
  author={Karimi, Ali and Faez, Karim and Nazari, Soheila},
  journal={IEEE Access}, 
  title={DEU-Net: Dual-Encoder U-Net for Automated Skin Lesion Segmentation}, 
  year={2023},
  volume={11},
  number={},
  pages={134804-134821},
  doi={10.1109/ACCESS.2023.3337528}
}

@INPROCEEDINGS{6610779,
  author={Mendonça, Teresa and Ferreira, Pedro M. and Marques, Jorge S. and Marcal, André R. S. and Rozeira, Jorge},
  booktitle={2013 35th Annual International Conference of the IEEE Engineering in Medicine and Biology Society (EMBC)}, 
  title={PH2 - A dermoscopic image database for research and benchmarking}, 
  year={2013},
  volume={},
  number={},
  pages={5437-5440},
  doi={10.1109/EMBC.2013.6610779}
}

@article{SIRINUKUNWATTANA2017489,
    title = {Gland segmentation in colon histology images: The glas challenge contest},
    journal = {Medical Image Analysis},
    volume = {35},
    pages = {489-502},
    year = {2017},
    issn = {1361-8415},
    doi = {https://doi.org/10.1016/j.media.2016.08.008},
    author = {Korsuk Sirinukunwattana and Josien P.W. Pluim and Hao Chen and Xiaojuan Qi and Pheng-Ann Heng and Yun Bo Guo and Li Yang Wang and Bogdan J. Matuszewski and Elia Bruni and Urko Sanchez and Anton Böhm and Olaf Ronneberger and Bassem Ben Cheikh and Daniel Racoceanu and Philipp Kainz and Michael Pfeiffer and Martin Urschler and David R.J. Snead and Nasir M. Rajpoot},
}

@article{7803544,
  author       = {Vijay Badrinarayanan and
                  Alex Kendall and
                  Roberto Cipolla},
  title        = {SegNet: {A} Deep Convolutional Encoder-Decoder Architecture for Image
                  Segmentation},
  journal      = {{IEEE} Trans. Pattern Anal. Mach. Intell.},
  volume       = {39},
  number       = {12},
  pages        = {2481--2495},
  year         = {2017},
  doi          = {10.1109/TPAMI.2016.2644615},
}

@inproceedings{bengio2007,
    author = {Bengio, Yoshua and Lamblin, Pascal and Popovici, Dan and Larochelle, Hugo},
    booktitle = {Advances in Neural Information Processing Systems},
    editor = {B. Sch\"{o}lkopf and J. Platt and T. Hoffman},
    pages = {},
    publisher = {MIT Press},
    title = {Greedy Layer-Wise Training of Deep Networks},
    volume = {19},
    year = {2006}
}

@article{larochelle2009,
    author = {Larochelle, Hugo and Bengio, Yoshua and Louradour, J\'{e}r\^{o}me and Lamblin, Pascal},
    title = {Exploring Strategies for Training Deep Neural Networks},
    year = {2009},
    issue_date = {12/1/2009},
    publisher = {JMLR.org},
    volume = {10},
    issn = {1532-4435},
    journal = {J. Mach. Learn. Res.},
    month = {jun},
    pages = {1–40},
    numpages = {40}
}

@inproceedings{9157032,
  author       = {Yassine Ouali and
                  C{\'{e}}line Hudelot and
                  Myriam Tami},
  title        = {Semi-Supervised Semantic Segmentation With Cross-Consistency Training},
  booktitle    = {2020 {IEEE/CVF} Conference on Computer Vision and Pattern Recognition,
                  {CVPR} 2020, Seattle, WA, USA, June 13-19, 2020},
  pages        = {12671--12681},
  publisher    = {Computer Vision Foundation / {IEEE}},
  year         = {2020},
  doi          = {10.1109/CVPR42600.2020.01269},
}

@article{LUO2022102517,
    title = {Semi-supervised medical image segmentation via uncertainty rectified pyramid consistency},
    journal = {Medical Image Analysis},
    volume = {80},
    pages = {102517},
    year = {2022},
    issn = {1361-8415},
    doi = {https://doi.org/10.1016/j.media.2022.102517},
    author = {Xiangde Luo and Guotai Wang and Wenjun Liao and Jieneng Chen and Tao Song and Yinan Chen and Shichuan Zhang and Dimitris N. Metaxas and Shaoting Zhang},
}

@article{Luo_2021, 
    title={Semi-supervised Medical Image Segmentation through Dual-task Consistency}, 
    volume={35}, 
    ISSN={2159-5399}, 
    DOI={10.1609/aaai.v35i10.17066}, 
    number={10}, 
    journal={Proceedings of the AAAI Conference on Artificial Intelligence}, 
    publisher={Association for the Advancement of Artificial Intelligence (AAAI)}, 
    author={Luo, Xiangde and Chen, Jieneng and Song, Tao and Wang, Guotai}, 
    year={2021}, 
    month=may, 
    pages={8801–8809} 
}

@inproceedings{kingma2014b,
    author = {Kingma, Diederik P. and Rezende, Danilo J. and Mohamed, Shakir and Welling, Max},
    title = {Semi-supervised learning with deep generative models},
    year = {2014},
    publisher = {MIT Press},
    address = {Cambridge, MA, USA},
    booktitle = {Proceedings of the 27th International Conference on Neural Information Processing Systems - Volume 2},
    pages = {3581–3589},
    numpages = {9},
    location = {Montreal, Canada},
    series = {NIPS'14}
}

@inproceedings{zhang2016,
    author = {Zhang, Yuting and Lee, Kibok and Lee, Honglak},
    title = {Augmenting supervised neural networks with unsupervised objectives for large-scale image classification},
    year = {2016},
    publisher = {JMLR.org},
    booktitle = {Proceedings of the 33rd International Conference on International Conference on Machine Learning - Volume 48},
    pages = {612–621},
    numpages = {10},
    location = {New York, NY, USA},
    series = {ICML'16}
}

@article{luo2021,
    author = {Liqun Luo },
    title = {Architectures of neuronal circuits},
    journal = {Science},
    volume = {373},
    number = {6559},
    pages = {eabg7285},
    year = {2021},
    doi = {10.1126/science.abg7285},
}

@inproceedings{10.1007/978-3-030-58526-6_45,
  author       = {Cheng Ouyang and
                  Carlo Biffi and
                  Chen Chen and
                  Turkay Kart and
                  Huaqi Qiu and
                  Daniel Rueckert},
  title        = {Self-supervision with Superpixels: Training Few-Shot Medical Image
                  Segmentation Without Annotation},
  booktitle    = {Computer Vision - {ECCV} 2020 - 16th European Conference, Glasgow,
                  UK, August 23-28, 2020, Proceedings, Part {XXIX}},
  series       = {Lecture Notes in Computer Science},
  volume       = {12374},
  pages        = {762--780},
  publisher    = {Springer},
  year         = {2020},
  doi          = {10.1007/978-3-030-58526-6\_45},
}

@article{XIONG2021101832,
    title = {A global benchmark of algorithms for segmenting the left atrium from late gadolinium-enhanced cardiac magnetic resonance imaging},
    journal = {Medical Image Analysis},
    volume = {67},
    pages = {101832},
    year = {2021},
    issn = {1361-8415},
    doi = {https://doi.org/10.1016/j.media.2020.101832},
    author = {Zhaohan Xiong and Qing Xia and Zhiqiang Hu and Ning Huang and Cheng Bian and Yefeng Zheng and Sulaiman Vesal and Nishant Ravikumar and Andreas Maier and Xin Yang and Pheng-Ann Heng and Dong Ni and Caizi Li and Qianqian Tong and Weixin Si and Elodie Puybareau and Younes Khoudli and Thierry Géraud and Chen Chen and Wenjia Bai and Daniel Rueckert and Lingchao Xu and Xiahai Zhuang and Xinzhe Luo and Shuman Jia and Maxime Sermesant and Yashu Liu and Kuanquan Wang and Davide Borra and Alessandro Masci and Cristiana Corsi and Coen {de Vente} and Mitko Veta and Rashed Karim and Chandrakanth Jayachandran Preetha and Sandy Engelhardt and Menyun Qiao and Yuanyuan Wang and Qian Tao and Marta Nuñez-Garcia and Oscar Camara and Nicolo Savioli and Pablo Lamata and Jichao Zhao},
}

@InProceedings{10.1007/978-3-030-32245-8_67,
    author="Yu, Lequan
    and Wang, Shujun
    and Li, Xiaomeng
    and Fu, Chi-Wing
    and Heng, Pheng-Ann",
    title="Uncertainty-Aware Self-ensembling Model for Semi-supervised 3D Left Atrium Segmentation",
    booktitle="Medical Image Computing and Computer Assisted Intervention -- MICCAI 2019",
    year="2019",
    publisher="Springer International Publishing",
    address="Cham",
    pages="605--613",
    isbn="978-3-030-32245-8"
}

@inproceedings{9577639,
  author       = {Xiaokang Chen and
                  Yuhui Yuan and
                  Gang Zeng and
                  Jingdong Wang},
  title        = {Semi-Supervised Semantic Segmentation With Cross Pseudo Supervision},
  booktitle    = {{IEEE} Conference on Computer Vision and Pattern Recognition, {CVPR}
                  2021, virtual, June 19-25, 2021},
  pages        = {2613--2622},
  publisher    = {Computer Vision Foundation / {IEEE}},
  year         = {2021},
  doi          = {10.1109/CVPR46437.2021.00264},
}

@misc{raffaele_mazziotti_2021_4488164,
  author       = {Raffaele Mazziotti and
                  Fabio Carrara and
                  Aurelia Viglione and
                  Lupori Leonardo and
                  Lo Verde Luca and
                  Benedetto Alessandro and
                  Ricci Giulia and
                  Sagona Giulia and
                  Amato Giuseppe and
                  Pizzorusso Tommaso},
  title        = {{Human and Mouse Eyes for Pupil Semantic 
                   Segmentation}},
  month        = feb,
  year         = 2021,
  publisher    = {Zenodo},
  version      = {1.0},
  doi          = {10.5281/zenodo.4488164}
}

@inproceedings{10376766,
  author       = {Yanfeng Zhou and
                  Jiaxing Huang and
                  Chenlong Wang and
                  Le Song and
                  Ge Yang},
  title        = {XNet: Wavelet-Based Low and High Frequency Fusion Networks for Fully-
                  and Semi-Supervised Semantic Segmentation of Biomedical Images},
  booktitle    = {{IEEE/CVF} International Conference on Computer Vision, {ICCV} 2023,
                  Paris, France, October 1-6, 2023},
  pages        = {21028--21039},
  publisher    = {{IEEE}},
  year         = {2023},
  doi          = {10.1109/ICCV51070.2023.01928},
}

@article{dosovitskiy2020,
  title={An image is worth 16x16 words: Transformers for image recognition at scale},
  author={Dosovitskiy, Alexey and Beyer, Lucas and Kolesnikov, Alexander and Weissenborn, Dirk and Zhai, Xiaohua and Unterthiner, Thomas and Dehghani, Mostafa and Minderer, Matthias and Heigold, Georg and Gelly, Sylvain and others},
  journal={arXiv preprint arXiv:2010.11929},
  year={2020}
}

@article{CIAMPI2022102500,
    title = {Learning to count biological structures with raters’ uncertainty},
    journal = {Medical Image Analysis},
    volume = {80},
    pages = {102500},
    year = {2022},
    issn = {1361-8415},
    doi = {https://doi.org/10.1016/j.media.2022.102500},
    author = {Luca Ciampi and Fabio Carrara and Valentino Totaro and Raffaele Mazziotti and Leonardo Lupori and Carlos Santiago and Giuseppe Amato and Tommaso Pizzorusso and Claudio Gennaro},
}

@inproceedings{hebbian_eccv_workshop,
  author       = {Luca Ciampi and
                  Gabriele Lagani and
                  Giuseppe Amato and
                  Fabrizio Falchi},
  title        = {A Biologically-Inspired Approach to Biomedical Image Segmentation},
  booktitle    = {Computer Vision - {ECCV} 2024 Workshops - Milan, Italy, September
                  29-October 4, 2024, Proceedings, Part {XIV}},
  series       = {Lecture Notes in Computer Science},
  volume       = {15636},
  pages        = {158--171},
  publisher    = {Springer},
  year         = {2024},
  doi          = {10.1007/978-3-031-91578-9\_10},
  bibsource    = {dblp computer science bibliography, https://dblp.org}
}

@inproceedings{DBLP:conf/nips/TarvainenV17,
  author       = {Antti Tarvainen and
                  Harri Valpola},
  title        = {Mean teachers are better role models: Weight-averaged consistency
                  targets improve semi-supervised deep learning results},
  booktitle    = {Advances in Neural Information Processing Systems 30: Annual Conference
                  on Neural Information Processing Systems 2017, December 4-9, 2017,
                  Long Beach, CA, {USA}},
  pages        = {1195--1204},
  year         = {2017},
}

@inproceedings{DBLP:conf/cvpr/IscenTAC19,
  author       = {Ahmet Iscen and
                  Giorgos Tolias and
                  Yannis Avrithis and
                  Ondrej Chum},
  title        = {Label Propagation for Deep Semi-Supervised Learning},
  booktitle    = {{IEEE} Conference on Computer Vision and Pattern Recognition, {CVPR}
                  2019, Long Beach, CA, USA, June 16-20, 2019},
  pages        = {5070--5079},
  publisher    = {Computer Vision Foundation / {IEEE}},
  year         = {2019},
  doi          = {10.1109/CVPR.2019.00521},
}

@article{DBLP:journals/nn/VermaKLKSBL22,
  author       = {Vikas Verma and
                  Kenji Kawaguchi and
                  Alex Lamb and
                  Juho Kannala and
                  Arno Solin and
                  Yoshua Bengio and
                  David Lopez{-}Paz},
  title        = {Interpolation consistency training for semi-supervised learning},
  journal      = {Neural Networks},
  volume       = {145},
  pages        = {90--106},
  year         = {2022},
  doi          = {10.1016/J.NEUNET.2021.10.008},
}

@article{chen2020b,
  title={Big self-supervised models are strong semi-supervised learners},
  author={Chen, Ting and Kornblith, Simon and Swersky, Kevin and Norouzi, Mohammad and Hinton, Geoffrey E},
  journal={Advances in neural information processing systems},
  volume={33},
  pages={22243--22255},
  year={2020}
}

@inproceedings{DBLP:conf/cvpr/VuJBCP19,
  author       = {Tuan{-}Hung Vu and
                  Himalaya Jain and
                  Maxime Bucher and
                  Matthieu Cord and
                  Patrick P{\'{e}}rez},
  title        = {{ADVENT:} Adversarial Entropy Minimization for Domain Adaptation in
                  Semantic Segmentation},
  booktitle    = {{IEEE} Conference on Computer Vision and Pattern Recognition, {CVPR}
                  2019, Long Beach, CA, USA, June 16-20, 2019},
  pages        = {2517--2526},
  publisher    = {Computer Vision Foundation / {IEEE}},
  year         = {2019},
  doi          = {10.1109/CVPR.2019.00262},
}

@inproceedings{DBLP:conf/eccv/OuyangBCKQR20,
  author       = {Cheng Ouyang and
                  Carlo Biffi and
                  Chen Chen and
                  Turkay Kart and
                  Huaqi Qiu and
                  Daniel Rueckert},
  title        = {Self-supervision with Superpixels: Training Few-Shot Medical Image
                  Segmentation Without Annotation},
  booktitle    = {Computer Vision - {ECCV} 2020 - 16th European Conference, Glasgow,
                  UK, August 23-28, 2020, Proceedings, Part {XXIX}},
  series       = {Lecture Notes in Computer Science},
  volume       = {12374},
  pages        = {762--780},
  publisher    = {Springer},
  year         = {2020},
  doi          = {10.1007/978-3-030-58526-6\_45},
}

@inproceedings{lee2013,
  title={Pseudo-label: The simple and efficient semi-supervised learning method for deep neural networks},
  author={Lee, Dong-Hyun and others},
  booktitle={Workshop on challenges in representation learning, ICML},
  volume={3},
  pages={896},
  year={2013}
}

@inproceedings{DBLP:conf/nips/HoJA20,
  author       = {Jonathan Ho and
                  Ajay Jain and
                  Pieter Abbeel},
  title        = {Denoising Diffusion Probabilistic Models},
  booktitle    = {Advances in Neural Information Processing Systems 33: Annual Conference
                  on Neural Information Processing Systems 2020, NeurIPS 2020, December
                  6-12, 2020, virtual},
  year         = {2020},
}

@inproceedings{DBLP:conf/nips/SohnBCZZRCKL20,
  author       = {Kihyuk Sohn and
                  David Berthelot and
                  Nicholas Carlini and
                  Zizhao Zhang and
                  Han Zhang and
                  Colin Raffel and
                  Ekin Dogus Cubuk and
                  Alexey Kurakin and
                  Chun{-}Liang Li},
  title        = {FixMatch: Simplifying Semi-Supervised Learning with Consistency and
                  Confidence},
  booktitle    = {Advances in Neural Information Processing Systems 33: Annual Conference
                  on Neural Information Processing Systems 2020, NeurIPS 2020, December
                  6-12, 2020, virtual},
  year         = {2020},
}

@article{DBLP:journals/nn/CiampiLAF26,
  author       = {Luca Ciampi and
                  Gabriele Lagani and
                  Giuseppe Amato and
                  Fabrizio Falchi},
  title        = {Biologically-inspired semi-supervised semantic segmentation for biomedical
                  imaging},
  journal      = {Neural Networks},
  volume       = {201},
  pages        = {108925},
  year         = {2026},
  doi          = {10.1016/J.NEUNET.2026.108925},
  bibsource    = {dblp computer science bibliography, https://dblp.org}
}

@article{DBLP:journals/artmed/LeeHWHLHT24,
  author       = {Chih{-}Kuo Lee and
                  Jhen{-}Wei Hong and
                  Chia{-}Ling Wu and
                  Jia{-}Ming Hou and
                  Yen{-}An Lin and
                  Kuan{-}Chih Huang and
                  Po{-}Hsuan Tseng},
  title        = {Real-time coronary artery segmentation in {CAG} images: {A} semi-supervised
                  deep learning strategy},
  journal      = {Artif. Intell. Medicine},
  volume       = {153},
  pages        = {102888},
  year         = {2024},
  doi          = {10.1016/J.ARTMED.2024.102888},
}

@article{DBLP:journals/artmed/MeiYZZCYW24,
  author       = {Chenyang Mei and
                  Xiaoguo Yang and
                  Mi Zhou and
                  Shaodan Zhang and
                  Hao Chen and
                  Xiaokai Yang and
                  Lei Wang},
  title        = {Semi-supervised image segmentation using a residual-driven mean teacher
                  and an exponential Dice loss},
  journal      = {Artif. Intell. Medicine},
  volume       = {148},
  pages        = {102757},
  year         = {2024},
  doi          = {10.1016/J.ARTMED.2023.102757},
}

@article{DBLP:journals/artmed/LiZYZWZLSW24,
  author       = {Jiajia Li and
                  Pingping Zhang and
                  Xia Yang and
                  Lei Zhu and
                  Teng Wang and
                  Ping Zhang and
                  Ruhan Liu and
                  Bin Sheng and
                  Kaixuan Wang},
  title        = {SSM-Net: Semi-supervised multi-task network for joint lesion segmentation
                  and classification from pancreatic {EUS} images},
  journal      = {Artif. Intell. Medicine},
  volume       = {154},
  pages        = {102919},
  year         = {2024},
  doi          = {10.1016/J.ARTMED.2024.102919},
}

@inproceedings{DBLP:conf/visapp/CiampiCAG22,
  author       = {Luca Ciampi and
                  Fabio Carrara and
                  Giuseppe Amato and
                  Claudio Gennaro},
  title        = {Counting or Localizing? Evaluating Cell Counting and Detection in
                  Microscopy Images},
  booktitle    = {Proceedings of the 17th International Joint Conference on Computer
                  Vision, Imaging and Computer Graphics Theory and Applications, {VISIGRAPP}
                  2022, Volume 4: VISAPP, Online Streaming, February 6-8, 2022},
  pages        = {887--897},
  publisher    = {{SCITEPRESS}},
  year         = {2022},
  doi          = {10.5220/0010923000003124},
  bibsource    = {dblp computer science bibliography, https://dblp.org}
}

@article{perez2025,
  title={Exploring scalable medical image encoders beyond text supervision},
  author={Perez-Garcia, Fernando and Sharma, Harshita and Bond-Taylor, Sam and Bouzid, Kenza and Salvatelli, Valentina and Ilse, Maximilian and Bannur, Shruthi and Castro, Daniel C and Schwaighofer, Anton and Lungren, Matthew P and others},
  journal={Nature Machine Intelligence},
  volume={7},
  number={1},
  pages={119--130},
  year={2025},
  publisher={Nature Publishing Group UK London}
}

@article{chen2024,
  title={Towards a general-purpose foundation model for computational pathology},
  author={Chen, Richard J and Ding, Tong and Lu, Ming Y and Williamson, Drew FK and Jaume, Guillaume and Song, Andrew H and Chen, Bowen and Zhang, Andrew and Shao, Daniel and Shaban, Muhammad and others},
  journal={Nature medicine},
  volume={30},
  number={3},
  pages={850--862},
  year={2024},
  publisher={Nature Publishing Group US New York}
}

@article{xu2024,
  title={A whole-slide foundation model for digital pathology from real-world data},
  author={Xu, Hanwen and Usuyama, Naoto and Bagga, Jaspreet and Zhang, Sheng and Rao, Rajesh and Naumann, Tristan and Wong, Cliff and Gero, Zelalem and Gonz{\'a}lez, Javier and Gu, Yu and others},
  journal={Nature},
  volume={630},
  number={8015},
  pages={181--188},
  year={2024},
  publisher={Nature Publishing Group UK London}
}

@inproceedings{DBLP:conf/iccv/CaronTMJMBJ21,
  author       = {Mathilde Caron and
                  Hugo Touvron and
                  Ishan Misra and
                  Herv{\'{e}} J{\'{e}}gou and
                  Julien Mairal and
                  Piotr Bojanowski and
                  Armand Joulin},
  title        = {Emerging Properties in Self-Supervised Vision Transformers},
  booktitle    = {2021 {IEEE/CVF} International Conference on Computer Vision, {ICCV}
                  2021, Montreal, QC, Canada, October 10-17, 2021},
  pages        = {9630--9640},
  publisher    = {{IEEE}},
  year         = {2021},
  doi          = {10.1109/ICCV48922.2021.00951},
  bibsource    = {dblp computer science bibliography, https://dblp.org}
}

@article{oquab2023,
  title={DINOv2: Learning Robust Visual Features without Supervision},
  author={Oquab, Maxime and Darcet, Timoth{\'e}e and Moutakanni, Th{\'e}o and Vo, Huy V and Szafraniec, Marc and Khalidov, Vasil and Fernandez, Pierre and HAZIZA, Daniel and Massa, Francisco and El-Nouby, Alaaeldin and others},
  year={2023},
  journal={Transactions on Machine Learning Research}
}

@inproceedings{DBLP:conf/midl/WuFFZYXL023,
  author       = {Junde Wu and
                  Rao Fu and
                  Huihui Fang and
                  Yu Zhang and
                  Yehui Yang and
                  Haoyi Xiong and
                  Huiying Liu and
                  Yanwu Xu},
  title        = {MedSegDiff: Medical Image Segmentation with Diffusion Probabilistic
                  Model},
  booktitle    = {Medical Imaging with Deep Learning, {MIDL} 2023, 10-12 July 2023,
                  Nashville, TN, {USA}},
  pages        = {1623--1639},
  year         = {2023},
  bibsource    = {dblp computer science bibliography, https://dblp.org}
}

@inproceedings{DBLP:conf/aaai/Wu0FXJ024,
  author       = {Junde Wu and
                  Wei Ji and
                  Huazhu Fu and
                  Min Xu and
                  Yueming Jin and
                  Yanwu Xu},
  title        = {MedSegDiff-V2: Diffusion-Based Medical Image Segmentation with Transformer},
  booktitle    = {Thirty-Eighth {AAAI} Conference on Artificial Intelligence, {AAAI}
                  2024, Thirty-Sixth Conference on Innovative Applications of Artificial
                  Intelligence, {IAAI} 2024, Fourteenth Symposium on Educational Advances
                  in Artificial Intelligence, {EAAI} 2024, February 20-27, 2024, Vancouver,
                  Canada},
  pages        = {6030--6038},
  year         = {2024},
  doi          = {10.1609/AAAI.V38I6.28418},
  bibsource    = {dblp computer science bibliography, https://dblp.org}
}

@inproceedings{DBLP:conf/miccai/ChenWS23,
  author       = {Tao Chen and
                  Chenhui Wang and
                  Hongming Shan},
  title        = {BerDiff: Conditional Bernoulli Diffusion Model for Medical Image Segmentation},
  booktitle    = {Medical Image Computing and Computer Assisted Intervention - {MICCAI}
                  2023 - 26th International Conference, Vancouver, BC, Canada, October
                  8-12, 2023, Proceedings, Part {IV}},
  pages        = {491--501},
  year         = {2023},
  doi          = {10.1007/978-3-031-43901-8\_47},
  bibsource    = {dblp computer science bibliography, https://dblp.org}
}
}

\clearpage
\appendix
\setcounter{page}{1}

\twocolumn[
\begin{center}
    {\LARGE\bfseries Supplementary Material}
\end{center}
\vspace{0.5cm}
]

\section{Experimental and Implementation Details}
\label{appendix:implementation}

In this section, we provide implementation details for the proposed method and the competing approaches. Our model is implemented using PyTorch, with all training and inference processes conducted on an NVIDIA DGX-A100. We also refer the reader to our repository available at \url{https://github.com/ciampluca/diffusion_semi_supervised_biomedical_image_segmentation}.

\subsection{Data Preprocessing and Evaluation Protocol}

To ensure fair comparisons, we followed a common experimental setting across all experiments. For the 2D datasets, data augmentation during training includes random vertical and horizontal flips as well as random rotations. Input images are resized to $128 \times 128$ for both training and inference. For the LA dataset~\cite{XIONG2021101832}, the same augmentation strategy is applied during training. Random patches of size $96 \times 96 \times 80$ are extracted for training, whereas inference is performed using a sliding-window approach with a 0.5 overlap ratio and patches of the same size.

We use a 10-fold cross-validation protocol for the 2D datasets and a 5-fold cross-validation protocol for LA, varying the training, validation, and test splits.

\subsection{Training and Optimization}

For the proposed method, training is performed using the Adam optimizer with an initial learning rate of $0.01$, momentum coefficients $\beta_1=0.9$ and $\beta_2=0.99$, and a weight decay of $5 \cdot 10^{-5}$. We use a multi-step learning rate scheduler that reduces the learning rate by a factor of 10 every 50 epochs. These hyperparameters were selected through a grid-search procedure based on validation performance.

We set the number of training epochs to 200. For the proposed method, the model snapshot obtained at the end of the unsupervised teacher-pretraining stage is used to initialize the teacher in the subsequent teacher--student co-training stage. Similarly, for two-stage baselines requiring unsupervised pretraining on the target dataset, the model snapshot from the final pretraining epoch is used to initialize the corresponding model for the second-stage training.

For the proposed method, the weight $\lambda_{\mathrm{u}}$ applied to the
unlabeled-data objective is increased linearly during training according to
\begin{equation}
\lambda_{\mathrm{u}}(e)
=
\lambda_{\mathrm{u}}^{\max}
\frac{e+1}{E},
\end{equation}
where $e\in\{0,\ldots,E-1\}$ denotes the current training epoch and $E$ is
the total number of training epochs. We set
$\lambda_{\mathrm{u}}^{\max}=1$.
The reconstruction and alignment terms introduced by the multi-round
extension share the same fixed weight, which is set to
$\lambda_{\mathrm{diff}}=1$.

For the competing methods, we use the SGD optimizer together with a multi-step learning rate scheduler to train the semi-supervised baselines and the second stage of the two-stage approaches, with RAD-DINO treated separately as described below. For two-stage approaches requiring unsupervised pretraining on the target dataset, we use the Adam optimizer during the first stage. Specifically, the initial learning rate is set to $0.5$ with SGD for the 2D datasets and to $0.1$ for the 3D LA dataset. For baseline pretraining performed with Adam, we use smaller initial learning rates of $0.001$ for the 2D datasets and $0.0001$ for LA. These optimization settings were selected through a grid-search procedure based on validation performance.

Additionally, for the weight $\lambda$ in the unsupervised loss functions of the competitor pseudo-labeling- and consistency-based methods, we increase $\lambda$ linearly with the number of epochs, following previous works, i.e.,
\begin{equation}
\lambda
=
\lambda_{\max}
\frac{\mathrm{epoch}}{\mathrm{max\_epoch}}.
\end{equation}
We set $\lambda_{\max}=5$, which was selected through a grid-search procedure based on validation performance.

\subsection{Two-stage Baseline Implementations}

Concerning the two-stage competitor approaches, we occasionally applied architectural modifications to better align each method's original task with the target task---semantic segmentation. Specifically, we added segmentation branches that were fine-tuned in the second stage and built on top of the features learned during the unsupervised stage.

\noindent \textbf{SuperpixSSL~\cite{10.1007/978-3-030-58526-6_45}}: no changes were necessary, as its architecture was already designed for semantic segmentation.

\noindent \textbf{RAD-DINO~\cite{perez2025}}: we initialized the model using the pretrained image encoder from the original work and kept it frozen during the second stage. Since RAD-DINO relies on a large-scale pretrained encoder, we do not retrain it from scratch on the comparatively small datasets considered in this work. Instead, we directly exploit the pretrained representation and train a task-specific segmentation decoder.
To produce segmentation maps, we added a decoding branch on top of the frozen RAD-DINO encoder. Each input image is mapped to a $37 \times 37 \times 768$ feature representation. The decoder consists of a sequence of transpose convolutions: first, a layer with 768 input and 256 output channels (kernel size 3, stride 1), followed by ReLU and BatchNorm; then a layer with 256 input and 128 output channels (kernel size 3, stride 2), followed by ReLU and BatchNorm; and a layer with 128 input and 64 output channels (kernel size 7, stride 3), also followed by ReLU and BatchNorm. The resulting features are upsampled through bilinear interpolation to $222 \times 222$ pixels and finally processed by a transpose convolution with 64 input channels and one output channel per class (kernel size 3, stride 1) to obtain the segmentation map.
This decoder configuration was selected through empirical evaluation of different architectural configurations. The decoder is trained independently on each dataset using the available labeled samples and the Adam optimizer with a learning rate of $0.01$, while the pretrained RAD-DINO encoder remains frozen.

\subsection{Diffusion Configuration}

Following standard denoising diffusion models~\cite{DBLP:conf/nips/HoJA20}, the cumulative signal-retention coefficient used in the main paper is defined as
\begin{equation}
\bar{\alpha}_t
=
\prod_{s=0}^{t}\alpha_s,
\qquad
\alpha_s = 1-\beta_s.
\end{equation}

The diffusion rates $\beta_s$ follow a linear schedule ranging from $\beta_0=0.0001$ to $\beta_{T-1}=0.02$. The total number of diffusion timesteps is set to $T=1000$.

Unless otherwise specified, we use $R=10$ stochastic reconstruction--segmentation rounds in the main experiments. To analyze the sensitivity of the proposed method to this parameter, the ablation study evaluates $R\in\{3,5,10\}$.

\end{document}